\documentclass[12pt]{article}

\usepackage{wqp}
\usepackage[nottoc,numbib]{tocbibind}
\usepackage{amsmath, amssymb, graphicx, xcolor, wrapfig, lipsum, bbm, url, subfig}
\usepackage{booktabs, multirow, tabularx}
\newcommand{\F}{\mathcal{F}}
\renewcommand{\L}{\mathcal{L}}
\newcommand{\M}{\mathcal{M}}
\newcommand{\D}{\mathcal{D}}
\newcommand{\R}{\mathbb{R}}
\newcommand{\x}{\mathbf{x}}
\newcommand{\z}{\mathbf{z}}
\newcommand{\blambda}{{\pmb{\lambda}}}
\newcommand{\prob}{\mathbb{P}}
\newcommand{\expectation}{\mathbb{E}}
\newcommand{\normal}{\mathcal{N}}

\DeclareMathOperator*{\argmax}{arg\,max}
\DeclareMathOperator*{\argmin}{arg\,min}

\newcommand{\alphad}{\textsf{AlphaD3M }}
\newcommand{\sklearn}{\textsf{scikit-learn }}
\newcommand{\evalml}{\textsf{EvalML }}
\newcommand{\tpot}{\textsf{TPOT }}
\newcommand{\autokeras}{\textsf{AutoKeras }}
\newcommand{\autoweka}{\textsf{auto-WEKA }}

\newcommand{\hyperopt}{\textsf{hyperopt }}

\newcommand{\mlbox}{\textsf{MLBox }}
\newcommand{\pycaret}{\textsf{PyCaret }}

\title{A Survey of Open Source Automation Tools for Data Science Predictions}
\author{Nicholas Hoell}

\affiliation{Research Technology \\WorldQuant Predictive}
\emailAdd{nicholas.hoell@wqpredictive.com}
\abstract{We present an expository overview of technical and cultural challenges to the development and adoption of automation at various stages in the data science prediction lifecycle, restricting focus to supervised learning with structured datasets.  In addition, we review popular open source Python tools implementing common solution patterns for the automation challenges and highlight gaps where we feel progress still demands to be made.}

\begin{document}
\maketitle

\listoffigures
\listoftables

\section{Overview}
\label{overview}
As the demand for robust predictive models has increased over the past decades, so too has the demand for tools data scientists can use to automate their workflow.  This can be advantageous for a number of reasons, including the following:

\begin{itemize}
\item {\bf Time savings: } Automated systems offer the potential for tremendous speed-up in a number of areas that data scientists spend their time.  Some common patterns that often have reusable components are data cleansing/standardizing, feature generation, and model tuning.  
\item {\bf Benchmarking: } The fast production of independent benchmarks to outperform and test against improves the quality of competitor models. 
\item {\bf Expanded userbase: } The more automation is built into data science solutions, the less specialized skills are required of users.  At the extreme, this offers the possibility of business users specifying a small number of high-level decisions and letting automation drive the subsequent pipelining (and, in some cases, even the reporting, e.g., \cite{automated_statistician}). 
\end{itemize}

While there are other important areas where automation has impacted the data science lifecycle, the focus of this survey is specifically on open source Python tools used for various steps in the automated machine learning pipeline for {\it supervised problems} pertaining to {\it structured data}.  A recent study \cite{adoption} that compared the adoption of automated tools for machine learning by organizational type found a plurality of respondents using automated tools only partially with significant variance by sector.  

This paper will present challenges faced in each of the three main stages of automation in a data science pipeline, together with an overview of some of the common concepts, terminology, algorithms, and open source implementations for addressing them. This is by no means meant to be considered an exhaustive study of all techniques (for this, we highly recommend the recent \cite{benchmark_survey} for AutoML and \cite{feature_selection} for feature engineering topics) or solutions but rather, its hopeful uniqueness lies in the following:

\begin{itemize}
\item The utility for the uninitiated reader to quickly gain an understanding of many of the algorithms being used in this active area
\item An overview of common open source Python tools available for practitioners and their potential trade-offs
\item A deeper understanding of common relevant design patterns, especially those concerning hyperparameter optimization
\item At-a-glance feature comparisons for a collection of AutoML tools
\item Extensive references for the reader looking to delve deeper
\end{itemize}

Our intended audience for this survey is data scientists with solid knowledge of major concepts in data science (clustering, linear models, etc) including some familiarity with deep learning, as well as a basic foundation in statistics.  We believe this survey can help to fill the current knowledge gap between data scientists primarily working in a business context and potentially lacking experience in AutoML concepts and packages and the daunting list of rapidly expanding open source options available.  As we pay particular attention to the utility of the packages and algorithms for the practicing data scientist, we also include discussion relevant to time series modelling tools. 

We will begin, in Section \ref{overview}, by introducing the three main components in a fully automated data science workflow, as first (we believe) described in \cite{prediction_factory}.  Then, in each of Sections \ref{prediction_engineering}, \ref{feature_engineering}, and \ref{model_engineering}, we will go in depth through each of these components, highlighting both popular algorithms as well as popular open source Python packages implementing them.  As the space of both algorithms and packages is quite large, not all will be covered and we refer the interested reader to \cite{compendium} for a well-maintained (as of this writing) compendium of model engineering packages.  

From a packages perspective, our survey will cover \textsf{sklearn}, \textsf{Auto-WEKA}, \textsf{auto-sklearn} and its variants, \textsf{MLBox}, \textsf{hyperopt-sklearn}, \textsf{EvalML}, \textsf{featuretools}, \textsf{autofeat}, \textsf{AutoKeras}, \textsf{AlphaD3M}, \textsf{AutoGluon-Tabular}, \textsf{Auto-PyTorch}, \textsf{TPOT}, \textsf{tsfresh}, \textsf{AutoTS}, \textsf{FLAML}, \textsf{Compose}, \textsf{Prophet}, \textsf{PyCaret},  \textsf{OptimalFlow}, \textsf{OBOE}, \textsf{GluonTS}, and \textsf{AdaNet}.

We present some open challenges remaining in Section \ref{remaining}.  For completeness, we include a self-contained primer on Bayesian optimization, including Gaussian Processes, in a closing Appendix as this plays a large role in  guiding the hyperparameter optimization and model selection process in several of the AutoML packages we discuss, and may not be something practicing data scientists have experience with.

\subsection{The main components}

There are several options for engineering automation in a data science pipeline, consisting of highly customizable levers.  We recommend \cite{prediction_factory} as a good overview of the general building blocks in such a pipeline, from which we shall borrow naming conventions in the list below.  

Given a dataset, $D$, consisting, potentially, of multiple tables, the building blocks underlying automated supervised learning are as follows\footnote{The terminology is taken from \cite{prediction_factory}}:

\begin{itemize}
\item {\bf Automated prediction engineering (autoPE):} This is the automated generation of {\it training examples} from $D$ producing tuples of inputs and their associated target labels.  This may additionally include the most recent relevant time at which the observation occurred as well as other potentially important inputs\footnote{This is slightly more general than the definition given in \cite{prediction_factory}.}. 

\item {\bf Automated feature engineering (autoFE):} This is the creation of new input variables associated to each training example.

\item {\bf Automated model engineering (autoME):} This is the creation of a model function, $m$, mapping feature vectors to predicted outcomes.  This stage is commonly thought of as  {\bf AutoML}, though in many systems AutoML includes both autoME and autoFE, as feature engineering can involve learning algorithms. 
\end{itemize}

\begin{figure}
\centering
\hspace*{-1.3cm}
\includegraphics[scale=.5]{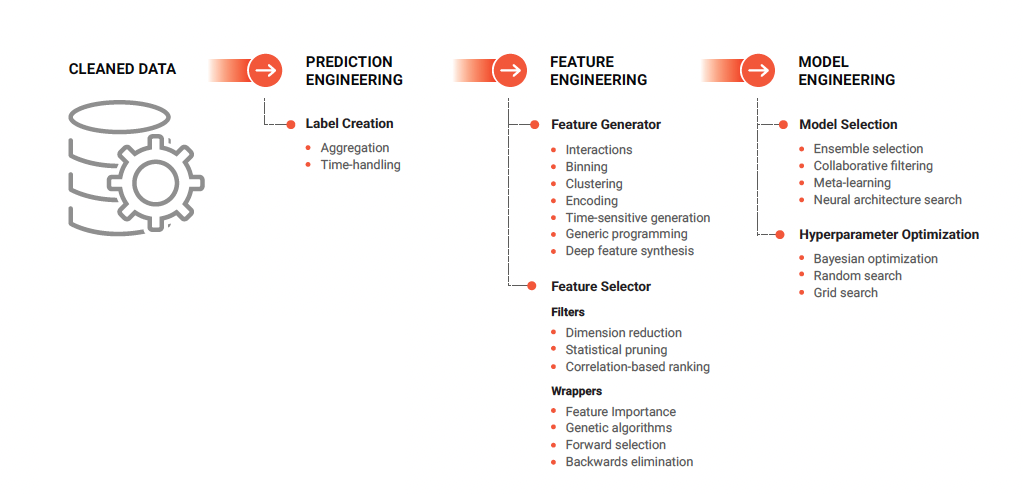}
\caption[Data science automation components]{Overview of the modular parts of the data science pipeline as first described in \cite{prediction_factory}, showing some of the internal levers/options for each component discussed in this article.}
\label{insight_factory}
\end{figure}

Each of these building blocks has its own substructure with its own unique challenges and common solutions.  We should note that the above presumes a specification of a prediction problem in the first place, which is generally proposed prior to modelling (see Section \ref{remaining} for more on this issue). 

After the autoPE stage, the dataset can be assumed to consist of labelled pairs $D = \{(\x^{(i)}, y^{(i)})\}_{i=1}^M$, with $\x^{(i)}$ the base feature vectors (assumed to be drawn from a data-generating distribution) and $y^{(i)}$ the associated target values or labels. The general data science prediction problem, generically, is to find a model $m$ for which

\[m(\x^{(i)}) \sim y^{(i)}, \qquad i=1, ..., M\]
approximately holds\footnote{The sense in which the relationship is approximate is determined by an appropriate loss function, measuring, essentially, goodness of fit on the entire dataset or batch thereof.}. In other words, the model should approximately capture the relationship between the distribution of target values and the data-generating distribution. 

One important stage in the overall data science pipeline is general data preprocessing which may consist of imputation of missing values, assessment and/or removal of outliers, normalization, handling duplicate observations, or others.  Strictly speaking, these tasks can be considered as data preparation preceding even the prediction engineering stage and for this reason, we omit a fuller discussion of these techniques other than to mention that many AutoML packages discussed below (for instance, \textsf{EvalML}, \textsf{MLBox}, \textsf{OptimalFlow}, and \textsf{auto-sklearn} to name a few) perform some of these preprocessing steps as part of the data preparation pipelining stages. To the author's knowledge no currently available open source Python AutoML tooling include advanced imputation strategies such as multivariate imputation by chained equations (MICE, see e.g. \cite{mice}), nearest-neighbour based imputation (see, e.g. \cite{knn}) or autoencoder based techniques (see, e.g. \cite{midas, midas_repo}), although some, for instance \textsf{PyCaret}, employ iterative imputation strategies\footnote{\textsf{PyCaret} uses Light Gradient Boosting Machine (LightGBM) algorithm.}.   As these preprocessing steps can consume a significant amount of practicing data scientists' time, this may be an area for impactful future developments.

As the autoPE process involves the least amount of technical complications and has the least open source libraries devoted to it, we briefly present an overview of it in the following before delving into the more involved autoFE and autoME in sections \label{feature_engineering} and \ref{model_engineering}, respectively.

\subsection{Automated prediction engineering}
\label{prediction_engineering}
In the article \cite{label}, the term {\it prediction engineering} was first introduced, formally, as 
\begin{quote}
``
the process that transforms time-driven relational data into feature vectors
and labels that can be used for machine learning"
\end{quote}

The authors then describe a general label-generating procedure based around two principles: a labeling function and a data traversal algorithm.  The labeling function's responsibility is to create labels from time slices of the dataset while the traversal algorithm orchestrates the aggregation, time-partitioning and application of the labelling function on appropriate subsets of the data. These ideas were implemented in the open source \textsf{Compose} package (see \cite{ compose_repo, compose_article}) by Innovation Labs at Alteryx.

Although the labelling process is automated by \textsf{Compose}, the implementation of it does require a bit more technical than some ``one-button" AutoML approaches as this step is responsible for a formal specification of the target instances.  To the authors' knowledge, \textsf{Compose} is the only open source Python tool currently dedicated to prediction engineering automation.  Additionally, currently available automated open source tooling does not seem to deal with automated stratification of samples or, more generally, processing that aims to find ``representative" examples from the data.  We feel this is an area for potential advances in the automated pipelining tools.

\section{Automated feature engineering}
\label{feature_engineering}
The space of available features for a given prediction task, we denote by $\L$, the so-called {\it feature landscape} or {\it feature space} for the given problem.  It is easy to see that even in the simplest case of a single target and a single base feature this space contains infinitely many elements. The exploration of $\L$ consists of successive application of feature transformations which result in a finite number of {\it derived features}, $\F\subset \L$, from which a final subset $\tilde{\F} \subseteq \F$ must be chosen. 

The feature engineering process thus consists of the following two stages:
\begin{enumerate}
\item {\bf Feature generation: } This stage consists of exploration of the feature landscape by construction of derived features, usually in an iterative manner.  The laws of combinatorics generally conspire to offer an explosion of derived features even for relatively few iterations.   This stage is often called {\it feature extraction}\footnote{This holds as well for techniques that combine generation and selection, such as PCA, as mentioned in the subsequent commentary.}. 
\item {\bf Feature selection: } This stage uses a fitness criteria to prune the newly-generated feature set $\D$ to a manageable size\footnote{We depart slightly from some traditional definitions, see e.g. \cite{polynomial}, in order to focus attention on different conceptual stages where possible.}.    This general problem, depending on exact specifications, is known to be NP-hard (see, e.g., \cite{min_subset, irrelevant_features}) and can be seen as a special case of approximating an optimal subset selection from $\F$'s $2^{\#\F}$ subsets.  Assuming the hyperparameter $\#\tilde{\F}$ to be fixed still yields a search problem among $\binom{\#\F}{\#\tilde{\F}}$ allowable feature subsets. 
\end{enumerate}

The boundary between these two steps can be blurred in practice.  For example, principal component analysis (PCA) is often used as a dimension reduction technique, constructing derived features that combine the input derived features and selecting those corresponding to the largest singular values of the empirical covariance matrix, and thus maximizing the variance of the original features in the lower dimensional space.  In other words, PCA often refers both to the generation of new features {\it and} to the selection of the new derived features.  While this line may be blurry in practice, the above division between generation and selection is a good, if not hard, cleaving of responsibilities in the feature engineering process. 

During an iterative feature generation cycle, the pruning process can take place ``mid-stream", i.e. concurrent with the feature generation, or ``downstream", i.e. after the full cycle of feature generation iterations has concluded\footnote{In practice there may be multiple iterations of the feature generation/selection cycle.}.  The advantage to placing the selection process downstream of the generation is the generation algorithm has the opportunity to explore more of the feature landscape that may not have been reached had feature pruning eliminated paths to the most informative/powerful features.   The trade-off is the enormous cost of storing the intermediate features required to generate the ultimately powerful end features. 

The advantage of using mid-stream selection algorithms is the potential speed-up in exploration and the avoidance of ``blind alleys" in the feature landscape.   This approach may not offer actual speed-up, depending on details of how the selection is implemented.  As well, these techniques suffer form the usual performance risks of greedy algorithms: by eliminating likely ``blind alleys" there is a risk of also eliminating paths through feature space that result in highly powerful features in the long run.  

In the following we outline common techniques in feature engineering and its automation in various Python packages.  Our coverage is far from complete but we hope the reader gets a sense of the main concepts and tricks of trade.  We also present, in Table \ref{feature_table}, the methods used by packages that perform automated feature engineering discussed in this survey.

\subsection{Generation techniques}
In the following we cover some of the common methodologies and open source implementations used in the automated generation of derived features for data science problems

\subsubsection{Elementary techniques}
We begin by very briefly covering a few of the more basic techniques for generating derived features, where possible highlighting open source Python implementations. 

\paragraph{Interactions.}
Feature {\it interactions} are obtained from simple products of existing features.  Here, we use the term more broadly and define {\it basic interactions} between features to be sums, differences, products, and ratios between numerical features.  The \pycaret library (see next section) offers automated generation of products and ratios of existing features as well as of multivariate polynomials generated by base features.  The \textsf{autofeat} and \textsf{featuretools} packages also offer a simple interface for quickly generating basic interaction features.

\paragraph{Encoding.}
One of the most common feature generation techniques is {\it one-hot encoding}.  This refers to transforming a categorical feature vector into a collection of bit vectors, each row of which containing a $1$ if and only if the given observation falls into the associated category.  One-hot encoding is part of the automated feature engineering stages of several packages including \textsf{EvalML}, \textsf{PyCaret}, and \textsf{auto-sklearn}.

\paragraph{Binning.} 
The \pycaret package automates the binning of continuous variables, converting them into categories on the basis of applying one dimensional $K$-means clustering, with number of clusters determined using {\it Sturges' rule} for optimal choice of bins.  The package also allows for automatically combining infrequent categories in features with large numbers of categorical values, as reducing the number of categories reduces the resultant number of features obtained when applying one-hot encoding.

\paragraph{Clustering.} Clustering refers to unsupervised partitioning of the dataset examples into a fixed number of clusters.  The derived feature from the clustering can be the categorical cluster label, the distance to the cluster representative (cluster centroid, say), or another derived quantity (see \cite{clustering} for a thorough overview).  A host of clustering techniques exist, with $k$-means (having an \sklearn implementation) and the density-based \textsf{HDBSCAN} (\cite{hdbscan_docs}) algorithms being two popular choices for data scientists.

\subsubsection{Some more sophisticated techniques}
Here we present a handful of techniques that go beyond some of the more basic techniques for feature generation discussed above and include some ideas or packages that may be novel to practicing data scientists. 

\paragraph{Genetic programming.}
Genetic programming is a common type of genetic algorithm (see, e.g. \cite{genetic} for a fuller description), a class of algorithm inspired by principles from evolution in the natural world which incorporates fitness selection, genetic recombination (sexual reproduction, although asexual techniques exist as well), and random mutation on a population of solutions to a given problem.  These solution populations undergo iterated evolution guiding them towards a superior solution.  

Genetic programming is a genetic algorithm whose populations are symbolic expression trees, with intermediate degree $k$ nodes given by primitive $k$-ary operators and with leaves given by symbols representing constants.  In the context of feature engineering, each expression tree represents a formula that can be applied to existing features to produce new derived features.  The recombination and mutation act on the feature space to produce a robust set of expressions with a fitness function selecting for an optimal property along the way (for instance, small linear correlations and non-redundant expressions,).  A popular open source implementation of genetic algorithms, including genetic programming is the \textsf{DEAP} package \cite{deap}., though we remark that \textsf{DEAP} may not be straightforward to implement into an automated pipeline as it is not particularly low-code. 

\paragraph{Deep feature synthesis.}
An algorithm called {\it deep feature synthesis} was presented in \cite{deep_feature_synthesis} which formed the basis for the \textsf{featuretools} Python library.  Deep feature synthesis was designed to work with data held in relational databases though reduced forms of it can work on single-table data.  The operators used to construct derived features are {\it aggregations} and {\it transformations}.  The former produces the {\it deep} part of deep feature synthesis by synthesizing data held in different tables via prescribed relationships among their unique keys and recursively traversing a graph made up of available primitive operations on existing features.  The latter consist of more tabular operations within a single table. 

The \textsf{featuretools} package implementing this algorithm offers very good graphical and even natural language representations of how the derived features are obtained, making it a particularly excellent choice for data scientists needing to speed up reporting.  The \pycaret package offers a somewhat related, though stripped down, version of synthesis by performing descriptive statistics on passed in groups of features as part of its feature engineering pipeline stage.

\paragraph{Synthetic features from random forests.}  In \cite{synthetic} the authors use random forests (see \cite{random_forests}) to create synthetic features to feed into a downstream forest.  The authors rely on interpreting the terminal node size hyperparameter of a random forest as a form of smoothness control and generate synthetic features by using the predicted values from a collection of random forests obtained with different node sizes.  For classification problems, these features are $\#classes$-dimensional vectors of estimated probability for each class, from each random forest.  The synthetic features are combined with the original base features to feed as input to a downstream random forest. This general paradigm of creating synthetic features, extended to include other classifier models, is used by the AutoML system \textsf{TPOT} and has a straightforward implementation using \textsf{scikit-learn}'s \textsf{ensemble} module as well.

\paragraph{Time series.}
There are often cases where datasets in question have a natural ordering to them, whether time-based or positional.  In the case where there in a natural one-dimensional ordering to the observational data, a host of common features present themselves as reasonable and informative.  We consider the case where each training example corresponds to various non-sequential features in addition to one or more time series.  An input example, the $i$'th say, may therefore take the form
\[ \x^{(i)} = (f^{(i)}_1, ..., f^{(i)}_n, \{X^{(i)}_1(t)\}_{t\in [t_0(X_1^{(i)}), t_{final}(X_1^{(i)}) ]}, ...,  \{X^{(i)}_m(t)\}_{t\in [t_0(X_m^{(i)}), t_{final}(X_m^{(i)}) ]}   ) \]
where in the above, the $f^{(i)}_j \in \mathbb{R}, j = 1, ..., n$ and $ \{X^{(i)}_k(t)\}_{t\in [t_0(X_k^{(i)}), t_{final}(X_k^{(i)})]}$ for $k=1, ..., m$ each represent time series, at possibly different cadence, length, and even type (categorical or numeric) associated to the observation $\x^{(i)}$. 
 In this case, the feature generation stage consists of ``flattening" the dataset by reducing each time series associated to an observation to a collection of single value, {\it unordered}, derived features.  

The \textsf{tsfresh} (Time Series FeatuRe Extraction on basis of Scalable Hypothesis tests, see \cite{tsfresh}) Python package is a common library for generating a host of traditional features for ordered datasets. It computes 794 different features from 63 characterization methods.  While deep feature synthesis can handle time series aggregations, the specificity of the methods used in \textsf{tsfresh}, including spectral features, make it extremely well suited for extracting informative derived features quickly. 

We close this subsection by referring the interested reader to the article \cite{time_series_review} which covers some of the additional automated feature generation tools available as part of automated forecasting libraries.  It presents a discussion of some of the general classes of features these tools can obtain.

\subsection{Filters, Wrappers, and All That}
As mentioned earlier, the process of selecting a good subset $\tilde{\F}$ of derived features from the $2^{\#\F}$ possible subsets of derived features can be NP-hard depending on the optimality criteria.   Although there are numerous possibilities for this search problem, two methods tend to occur in practice more often than others (see, e.g. \cite{filter_wrapper}): {\it filters} and {\it wrappers}.  Filter methods are those that use only properties intrinsic to the data, i.e. they are model-independent.   These are often considered a form of preprocessing as they do not involve the predictive model at all. Wrapper methods rely on using a machine learning model to evaluate the appropriateness of a proposed derived feature subset (see \cite{wrappers}). In addition, there are hybrid methods that combine filter and wrapper (see, e.g. \cite{paretoset}).  A third class is {\it embedded} techniques in which the selection process is incorporated directly into the modelling, for instance $l_1$-regularized regression models.  In this article we only focus on modular components and therefore restrict our attention to the filter and wrapper techniques below. 

The articles \cite{wiki_feature_selection} and \cite{stability} give  good, short surveys of some of the main concepts and techniques in the field.  Our presentation in the below is far from exhaustive; we aim to present a few popular options for the feature selection stage with open source Python implementations. Interested readers may also find \cite{variable} particularly valuable.

\subsubsection{Filters}
In the following, we cover some of the main ideas behind common approaches to filter feature selection algorithms as well as discuss some of the open source packages that offer user-friendly implementations of them.  Some well-known filtering techniques we omit for brevity are FOCUS, Relief, and the method of Cardie (see \cite{cardie}), each of which is discussed in \cite{irrelevant_features}.  Our presentation divides the techniques into three non-exclusive categories: those based on classical dimension reduction, those based heavily on information or correlation among features, and those based on statistical methods.  While these categories are somewhat artificial, we hope it clarifies some of the differences in rationale behind the approaches.

\paragraph{Dimension reduction.}
While traditional dimension reduction techniques were often not aimed at increasing the discriminatory power of downstream models, they have become an increasingly important part of the feature engineering stages of data science pipelines.  These methods combine both feature generation and selection; the input feature vectors are not simply reduced in quantity, they are transformed as well.  The transformation and selection are fundamentally entangled and not independently modularizable in a straightforward way.

The {\bf mutual information}, $I(X;Y)$, between two random variables $X$ and $Y$, with distribution functions $p_X$, $p_Y$, respectively, and joint distribution $p_{XY}$, is given by 
\[ \mathbb{E} \log (\frac{p_{XY}}{p_X p_Y})\]
which measures the dependency of variables $X$, $Y$.  In the above, the expectation is taken with respect to the joint distribution of $(X,Y)$. Among other properties, it is straightforward to show that $I(X;Y)$ is zero if and only if the variables $X$ and $Y$ are statistically independent.  Information-based filtering are techniques that, at their core, aim to reduce features based on higher mutual information.  Mutual information can be estimated using the \textsf{scikit-learn} package which is based on algorithms discussed in \cite{mutual_information, mutual_info} on entropy estimations.  Mutual information has a well-known bias in favour of larger numbers of values, see, e.g. \cite{quinlan, hall_thesis}.

{\bf Principal feature analysis} \cite{pfa} is a technique to select a subset of the derived features with low mutual information.   The method is based on the observation that the absolute value vectors produced by taking the rows of the (truncated) eigenvector matrix obtained from the eigendecomposition of the covariance matrix are in close Euclidean proximity for features with high mutual information. Based on this observation, $k$-means clustering in the lower-dimensional space produced from the row span of these absolute value vectors will find clusters with similar informational content, and the original features corresponding to the best representatives from each cluster will be the ultimately selected features. While the author could not find an open source Python package implementing principal feature analysis, this algorithm is a straightforward chaining of PCA and clustering found within the \textsf{scikit-learn} package.  Of course, a clear disadvantage of a straightforward implementation is the dependence on the nondeterministic clustering. 

While mutual information can directly measure statistical independence, correlations can, at times, be a reasonable proxy (as zero correlation is a necessary condition of independence and a sufficient condition for {\it linear} independence) and thus falls into this class of filtering.  This is highly intuitive: the smallest set of feature vectors spanning a $d$-dimensional subspace of the feature space will necessarily be linearly independent and, therefore, are uncorrelated.  PCA, as discussed in the previous section, falls into this category of selection as the principal components are orthonormal eigenvectors of the empirical covariance matrix and so are uncorrelated.  PCA is automated in the \pycaret package (see next section) as part of its feature engineering stage. 

In \cite{kernel_pca} a generalization of PCA, {\bf kernel PCA}, is introduced in which base features are embedded into a (higher-dimensional) new feature space in which eigenproblems for the empirical covariance matrix in the embedded space can be solved to get projections onto dominant eigenvectors, as in usual PCA (see also \cite{kernel_pca_features}).  The high dimensional embedding allows for the process to efficiently handle potential nonlinearities in the dataset as {\it explicit} embeddings are not actually required to compute the projections, only inner products among embedded vectors are required for the procedure.  As with linear PCA, projections can be truncated to projections on the most dominant (in terms of sizes of eigenvalues) eigenvectors.  Both PCA and kernel PCA rely on covariance matrices and all techniques reliant on covariance estimation can be sensitive to outliers (see \cite{outliers} and references therein). Kernel PCA is automated as part of preprocessing in some AutoML packages, such as \textsf{Auto-PyTorch} (see Section \ref{model_engineering}). 

PCA and kernel PCA are often viewed not just as feature engineering steps but also as dimension reduction techniques and derive their power essentially from scale imbalances in the distribution of points along suitable coordinates.  These differences in scale imply a reasonable approximation of the data should exist in a lower-dimensional space.  A different approach, based on the idea that the true dimension of the dataset is lower than the number of features, is given by locally linear embedding (LLE) \cite{lle, lle_overview}.  In LLE, data from an $n$-dimensional space is assumed to be drawn from a distribution of points on a $d < n$ dimensional manifold.  Because smooth manifolds are locally flat (see \cite{spivak}), the low-dimensional manifold hypothesis of LLE implies that a suitable linear representation exists of data points in terms of their near neighbours.  Once this linear map is obtained, projections to lower dimensions follow by minimizing an objective that penalizes variation from the linear relationship in the high-dimensional space holding in the lower-dimensional space, enforcing as much of the local geometry to hold in the lower dimensional space as possible (see also \cite{lle_score} for another take on using LLE in feature selection). Both kernel PCA and LLE have straightforward implementation in \textsf{sklearn}. 

Finally, Independent Component Analysis (ICA) is a technique for uncovering latent, non-Gaussian, independent, features from a dataset. Namely, observations are assumed to be generated by independent, non-Gaussian components, viz. $\x^{(i)} = A\mathbf{s}^{(i)}$ where $A$ is a mixing matrix and the entries of $\mathbf{s}^{(i)}$ represent statistically independent latent components. The independent components can be approximately determined by a projection pursuit algorithm, \textsf{FactICA} (introduced in \cite{ica}), that finds directions to project $\x^{(i)}$ onto that result in maximizing the kurtosis of the result\footnote{Recall that for a random variable $X$, the kurtosis can be defined by $\expectation X^4 - 3 (\expectation X^2)^2$ and is zero for normal distributions (see \cite{ica}).}.  In \cite{kurtosis}, the authors rank-order the induced latent features obtained from ICA by their kurtosis to obtain a feature subset.  As with other approaches in this section, this approach combines both generation and selection. ICA is built into the automated preprocessing portion of the \textsf{auto-sklearn} package.

\paragraph{Information and correlation.}
In the below, we outline a few methods for filtering that use information explicitly, or use correlation as a proxy.

A somewhat related method, not reliant upon statistical testing, is to rank features by their correlation (or covariance) with the target variable.  In the case of a univariate linear least squares regression model this gives the square root of the coefficient of determination, namely the fraction of the variance in the response explained by the variance in the given feature.  Extensions of this approach, and its relation to traditional statistical tests, is further discussed in \cite{variable}.

Another simple example of this kind of filtering is the removal of all features whose correlation with other features exceeds a threshold.  Linear correlations can easily be estimated from the values of the features and do not require direct distribution estimation.  This property makes them a desirable proxy for information between random variables in some instances. A related issue is multicollinearity, wherein one feature may have linear dependence on possibly several others.  The package \textsf{PyCaret}, for example, automates the handling of multicollinear features as part of the overall AutoML pipeline.

The article \cite{paretoset} introduces a hybrid approach whose filter stage uses a unique multi-objective optimization to prune features. For a given feature subset $\tilde{\F} \subset \F$, three objectives are calculated: the (negative of) average mutual information between features in $\tilde{\F}$ and target $Y$, the average mutual information between features in $\tilde{\F}$, and $\#\tilde{\F}$. The {\bf Pareto frontier}\footnote{Due to the classic analogy of optimizing for location and price of coastal hotels, this frontier is commonly called a {\it skyline} (see, e.g., the seminal article \cite{skyline}). 
} for a multi-objective optimization problem is a set of Pareto efficient (i.e. non-dominated\footnote{In this context, these are solutions for which no subset of features performs at least as well on 2 objectives and better on the remaining one. }) solutions.  The authors use a forward selection strategy based on finding an approximation of the Pareto frontier by growing the size of examined subsets while finding subsets that are non-dominated according to the first two (information-based) objectives.  To the authors knowledge, Pareto optimization is not performed in any open source automated tooling, however the \textsf{DEAP} package includes implementations of popular multi-objective optimizations such as NSGA-II (see \cite{nsga}).

In the doctoral thesis \cite{hall_thesis}, Mark Hall outlines a method for feature selection, {\bf correlation-based feature selection} (CFS), for classification problems, based on the following idea:

\begin{quote}
``A good feature subset is one that contains features highly correlated with
(predictive of) the class, yet uncorrelated with (not predictive of) each other"
\end{quote}
Stated this way, CFS aims to find a trade-off between {\it relevancy} (dependence of the target on the response, as above) and {\it redundancy} (correlations among the derived features)\footnote{The article \cite{variable} provides several helpful cautionary tales about naive application of ranking schemes. Among other things, they show that identically distributed variates may not be immediately redundant, irrelevant features may combine to give relevance as a group, and high correlation does not imply uselessness in separating classes.}\footnote{As discussed in \cite{wrappers}, some machine learning algorithms have highly  robust performance in the face of irrelevant variables but degrade when given highly correlated (redundant) relevant features.}. Features are retained based on their high relevance, low redundancy, and ability to predict classes in areas of the features space that are not predicted well by existing features.  For a classification prediction on targets $C$, a proposed subset, $\tilde{\F}\subseteq \F$ is scored according to 
\begin{equation}
\label{cfs}
 s(\tilde{\F}) \doteq  \frac{\sum_{f\in \tilde{\F}}\rho_{cf} } {\sqrt{\#\tilde{\F} + 2 \sum_{f, g \in \tilde{\F}} \rho_{fg}}} 
 \end{equation}
where $\rho_{fg}$ is the correlation coefficient between features $f,$ and $g$, and $\rho_{cf}$ is the correlation coefficient\footnote{These coefficients are not the usual linear Pearson coefficients, but represent a more general sense of the word ``correlation".  In Hall's presentation the coefficients can be symmetrical versions of relief, minimum description length, or symmetric uncertainty. See \cite{hall_thesis} for full details.} between feature $f$ and target values $c$.  The numerator in \eqref{cfs} indicates how predictive the features are, while the denominator penalizes  non-parsimony and redundancy. The \textsf{pyautoweka} package, which implements the Java \textsf{Auto-WEKA} application \cite{pyautoweka, autoweka} (described in more detail below)  performs CFS in addition to informational and entropy-based filtering as part of its general automated feature and model selection algorithm.

\paragraph{Statistical Pruning.}
A related technique to the informational approaches above are {\it statistical filtering} techniques.  The popular \textsf{scikit-learn} package (see \cite{sklearn, sklearn_api}) offers a host of basic statistical feature selection techniques through its \textsf{feature\_selection} module \cite{sklearn}.   These include methods such as \textsf{SelectKBest} which uses scores based on $p$-values of passed hypothesis tests of the features, \textsf{VarianceThreshold}, which makes sure that variance in a feature exceeds a minimal limit and removes those features below, as well as others.  Some of these methods are  automated in the \textsf{OptimalFlow}, \evalml and \tpot (see next section)  feature selection pipeline stage., while some implement their own takes on these patterns (such as \textsf{MLBox}'s automated implementation of variance thresholding).

A very good example of automated statistical filtering is outlined in \cite{tsfresh2}, as implemented in \textsf{tsfresh},  wherein for each derived feature, a $p$-value is computed to determine whether the derived feature is {\it relevant} for the prediction of the target. Here, {\it irrelevance} is equivalent to zero mutual information between the given feature and the target.  Each univariate feature has its $p$-value chosen (using potentially different hypothesis tests) resulting in a vector of $p$-values.  From here, a Benjamini-Yekutieli procedure is applied to avoid false positives (meaningless features making it through this round of filtering) among these $p$-values.  This procedure is based on a method for multiple comparison testing that controls the so called false discovery rate, $FDR$, defined by
\[ FDR = \mathbb{E}[\frac{\#\{\text{incorrect rejections of null hypothesis}\}}{{\#\{\text{rejections of null hypothesis}\}}}]\]
Feature reduction then takes place based on the $p$-values while keeping bounds on the $FDR$.  Of course, other techniques, like the simple Bonferroni correction or other multiple comparison solutions may be adopted in place of $FDR$-controlled methods.  In addition to the above, there can be overlap with the information pruning techniques mentioned previously when including the target variable in the information pruning basket.

\paragraph{A final remark on relevance.}
We close this discussion of filtering by noting that feature relevance is not at all an easy determination as there are many different, incompatible, definitions of relevance in the literature.  For examples of these definitions, see Kohavi and John's article \cite{wrappers} from which we quote:
\begin{quote}
``...it is important to realize
that relevance according to these definitions does not imply membership in the optimal
feature subset, and that irrelevance does not imply that a feature cannot be in the optimal
feature subset."\footnote{Optimality is given a precise meaning in their paper.}
\end{quote}\

\subsubsection{Wrappers}

In general, wrapper methods treat the predictive model as a black box, and assess fitness of a subset of features by looking at performance on that subset.  Although there are many approaches to wrappers (for instance, type of model, whether evaluation uses cross-validation or holdout, etc) many of the methods fall into two types: {\it forward selection} and {\it backward elimination}.   Forward selection wrappers begin with an initial set of features (the empty set, say) and proceed to add in features one at a time, while backward elimination rests on winnowing down the initial set of features by removing redundant or irrelevant features.  Basic implementations of this wrapper paradigm are implemented in the \sklearn \textsf{model\_selection} module. Of course, the search problem can be carried out with different algorithms and there are reasonable reasons for each approach (see \cite{wrappers, variable} for in-depth discussions on this topic).

\paragraph{Feature importance.}  The term {\it feature importance} can refer to a handful of different techniques.  A simple case is the case of linear models, where model coefficients can be viewed as feature importances when the features are appropriately normalized.  In the classic article \cite{random_forests}, permutation feature importances are explored in the context of random forests.  This amounts to looking at the resultant decrease in performance (misclassification rate, say) when each feature undergoes a random permutation of its values, keep all other features fixed.  \textsf{EvalML} (see next section) and \sklearn (which can be used with non-native variants such as \textsf{XGBoost}) include this form of importance as part of their basic suites. 

Sticking with forests, the determination of splits, the Gini index\footnote{If $\pi_i(t)$ is the estimated probability of class $i$ in node $t$, $\sum_{i \leq \# classes}\pi_i(t)(1-\pi_i(t))$ denotes the Gini index for $t$, see \cite{hastie}.} (or impurity) is often used for classification problems and sum of squares for regression (see, e.g. \cite{impurity} for more details). \textsf{XGBoost} offers multiple additional options for measuring feature importances, such as the number of times a given feature was used to provide a splitting of the data across all trees in the forest as well as the gain from each feature used in model construction/pruning, among others (see \cite{xgboost} for full details). Fully automated random forest impurity based feature importance is implemented by \textsf{MLBox}, \textsf{EvalML}, and other autoML packages.  

Building on work in \cite{ranking} for the case of filters, in \cite{boruta} a wrapper algorithm and R package, \textsf{Boruta},  are presented which are based around random forest feature importances and meant to find all relevant features for a given classification problem. Each feature in the dataset is used to create a ``shadow" feature whose values are permutations of the base feature.  Then, the enlarged dataset with the original and shadow features is used to perform a prediction and feature importances are computed.  The maximum $Z$ score is calculated among all shadow attributes (MZSA)\footnote{The $Z$ score of a variable $X$ is $\frac{X-\expectation X}{\sqrt{var(X)}}$.  In this case, the importances are random variables.} and those features with $Z$ scores testing significantly less than the MZSA are removed, while those testing significantly above are counted as important.  This process repeats until a stopping criteria is met. \textsf{Boruta} has an open source Python implementation (see \cite{boruta_py_repo}) as well as a Python extension \textsf{BoostARoota} \cite{boostaroota} which uses a variant of the Boruta algorithm with \textsf{XGBoost} for increased speed relative to the traditional random forest based \textsf{Boruta}. \textsf{Boruta} is part of \textsf{PyCaret}'s automated feature selection module. 

{\bf Shapley additive explanations} (SHAP, see \cite{shap}) can provide an alternative feature importance paradigm. Traditional Shapley values can be defined by 
\begin{equation}
\label{shapley}
\phi(f_i) = \sum_{S \subset \F \setminus \{f_i\}} c(S)(m_{S\cup \{f_i\}}(\x_{S\cup \{f_i\}}) - m_S(\x_S) )
\end{equation}
where $f_i$ is the $i$'th feature, $c(S) = \frac{\#S!(\#\F - \#S - 1)!}{\#\F!}$, $m_B$ denotes the predictive model obtained by training on the subset $B\subset \F$ features and $\x_B$ the corresponding inputs from the feature subset, $B$.  Morally speaking, the SHAP values are a variation of \eqref{shapley} wherein $m_B(\x_B)$ is approximated by a conditional expectation $\expectation[m(z)\mid z_B]$ where $z_B$ is a binary vector indicating which features are selected, and $m$ is the full predictive model.  As the sum in \eqref{shapley} extends over subsets of a feature set, sampling can be performed to give a tractable approximate solution, among other approaches (see \cite{shap} and \cite{shap_tutorial} for more details).  SHAP values are implemented in, e.g., \textsf{EvalML} as part of model explainability.

\paragraph{Genetic algorithms.}
A genetic algorithm based wrapper technique is to use a performance based fitness function during the evolutionary cycle.  A clear example of this general idea in the context of regression problems with categorical features, using the \textsf{DEAP} package, is found in \cite{genetic_kaggle}.  Similar ideas were explored in \cite{genetic_feature}. 

In \cite{scaling_tree_based}, an additional feature subset selector pipeline was added to the basic \tpot package (see Section \ref{genetic_tpot} for more details on the \tpot package) that performs removal of input features as a pipeline module.  The \tpot evolved model then can be used as a wrapper to select the best performing feature subset.  Expanding on the \tpot framework, in the article \cite{biomedical_tree_based_pipwlinw} a method inspired from genetics is added, paring down the full feature set to a restricted number of parameterized feature pairs. Each pair is then assessed using its accuracy on a custom tree built just on those two features.  This method  presumes the possible strong interaction between two features as can happen between two genes.

\paragraph{LASSO.}
The least absolute shrinkage and selection operator (LASSO) \cite{lasso_1, lasso_2} is an $\ell^1$-regularized variant on the linear least squares regression problem, that tries to solve
\begin{equation}
\label{lasso}
\argmin_{\pmb{\theta}} ||\begin{bmatrix} 1 & (\x^{(1)})^T  \\ &\vdots \\ 1& (\x^{(m)})^T  \end{bmatrix} \pmb{\theta} - \mathbf{y}|| + \lambda \sum_{i\leq n+1}|\theta_i|
\end{equation}
The first term in the above is a {\it fidelity} term penalizing poorness of fit to a linear mode, while the second term helps promote sparsity of the resultant $\pmb{\theta}$; the gradient of \eqref{lasso} will produce terms like $\frac{\partial}{\partial \theta_i}|\theta_i| = \frac{|\theta_i|}{\theta_i} = sgn(\theta_i)$ when $\theta_i \neq 0$ so that gradient descent will result in components of $\pmb{\theta}$ not improving fidelity tending to zero.  LASSO on its own, as a predictive model, is an embedded technique rather than a wrapper, but it can be repurposed as part of a larger wrapper method. 

The \textsf{autofeat} Python package \cite{autofeat} offers a multi-step feature reduction technique, consisting of removal of highly correlated features, followed by division of data into chunks on which $\ell^1$-regularized models are trained to guide feature selection.  This process repeats through several rounds of subsampling.  In addition, the package implements a form of \textbf{symbolic pruning},  utilizing the \textsf{SymPy} library to simplify and reduce redundant symbolic expressions.  This eliminates redundant but complex features without evaluating their associated expressions.  The popular autoML package \mlbox also offers automated LASSO feature selection as part of its preprocessing pipeline optimization.

%
%

\begin{table}[htp]
\centering
\resizebox{\textwidth}{!}{

    \begin{tabular}{  l p{2.1cm}  p{8cm}  p{8cm} }
        \toprule
\textbf{Package}      
& \textbf{Low-Code}   
& \textbf{Preprocessing}
& \textbf{Feature engineering} \\\midrule

\textsf{MLBox}&Yes&Cleaning, duplicate handling, typecasting, imputation, encoding & Variance thresholding, $L1$, random forest feature importances\\

\textsf{OptimalFlow}&Yes&Imputation, standardization, outlier winsorizing, encoding&Backward elimination, statistical pruning\\

\textsf{hyperopt-sklearn}&No&Encoding, scaling, time series lag selection, tf-idf, restricted Boltzmann machines &PCA, column $K$-means\\

\textsf{AutoGluon-Tabular}&Yes& Encoding, datetime handling, imputation, duplicate handling & None\\

\textsf{AlphaD3M}&Yes&Imputation, encoding, scaling&PCA, tf-idf, variance thresholding, statistical pruning\\

\textsf{Prophet}&Yes&None&None\\

\textsf{PyCaret}& Yes& Imputation, encoding, SMOTE, outlier removal&Interactions, trigonometric features, binning, clustering, permutation importance, correlation with target, removal of collinear features, PCA, Boruta\\

\textsf{EvalML}&Yes&None&None\\

\textsf{featuretools}&Yes&Data typing & Deep feature synthesis.  Customizable aggregations and transformations.  Basic selection based on low information or constant as well as removal of highly correlated features\\

\textsf{OBOE}&Yes&Imputation, encoding, standardization&None\\

\textsf{TPOT}&Yes&Scaling, encoding, normalization&Binning, FastICA, synthetic feature generation, feature agglomeration, PCA, polynomial features, forward selection, statistical pruning, variance thresholding, elimination\\

\textsf{AdaNet}&No&None&Implicit neural features\\

\textsf{FLAML}&Yes&Encoding, scaling, class balancing, stratified sampling&None\\

\textsf{tsfresh}&Yes&Imputation&Statistical pruning (false discovery rate minimization)\\

\textsf{AutoTS}&YEs& Cleaning and shaping, outlier handling, detrending, shifting, filtering, transforming& Not pipeline-indpendent\\

\textsf{Auto-Keras}&Yes&Typecasting, encoding& Implicit neural features\\

\textsf{Auto-WEKA} family&Yes&None&Forward selection, CFS subset selection, information gain evaluation, principal components evaluation, symmetrical uncertainty, gain ratio, Pearson correlation, 1-R\\

\textsf{auto-sklearn} family &Yes& Encoding, imputation, class balancing, rescaling& Extremely random trees preprocessing, fastICA, feature agglomeration, kernel PCA, random kitchen sinks, linear SVM preprocessing, nystroem sampler, PCA, polynomials, random trees embedding, percentile selection, select rates \\

\textsf{autofeat}& Yes& Encoding& Buckingham Pi theorem, interactions, recursive application of combinations of ratios, sums, trigonometric functions, absolute value and others.  Wrapper selection based on higher performance over corrupted version. \\

        \bottomrule
    \end{tabular}}
    
   \caption[Automated feature engineering components in various packages.]{Feature engineering and preprocessing comparison of packages discussed in this survey. For the purposes of this survey a package is deemed ``Low-Code" if it demands a {\it very low} learning curve from a user already familiar with \textsf{scikit-learn}, which is an industry standard for non-automated machine learning. }
    \label{feature_table}
\end{table}

\paragraph{Stability.} A distinct kind of wrapper technique is one based on {\it stability}, which can be interpreted in a number of different ways\footnote{In the literature, stability generally refers to what we call here {\it dataset stability}.  We use the term {\it stability} in a looser, more general, way.}. In \cite{paretoset}, the authors use an approach based on {\it classification stability}: for each subset $\tilde{\F}\subset \F$ of the feature set, two scores is computed based on classification rates for a handful of classifiers. The first is $A(\tilde{\F}) = \max_{f \in \M}R(f, \tilde{\F}) -  \min_{f \in \M}R(f, \tilde{\F})$, where $\M$ is a set of classifiers and $R(f, \tilde{\F}))$ represents the rate of correct classification with model $f$ using feature subset $\tilde{\F}$, and measures the overall worst variation in performance with a given subset.  The other objective is simply the average $R(f, \tilde{\F})$ over the classifier set.  Subsets can be retained based on minimization of the former and maximization of the latter objectives.  This multi-objective optimization can help ensure both reliability and performance of the final feature subset on a wide range of models.

In \cite{stability_index}, a modification to forward selection is proposed, which we will refer to as {\it dataset stability}.  The algorithm rests on the fact that error rates are random variables, as they depend on the data-generating distribution and therefore are influenced by sampling.  Since wrappers generally assess feature subsets on the basis of these random variables (error rates), the {\it sequence} of feature sets obtained during a single run of forward selection, is also a random variable, and ultimately dependent upon the training set used in the selection process. The proposed algorithm works by performing $K$ runs of forward selection, each producing a sequence of obtained feature sets, 
\[S_i = \tilde{\F}^{(i)}_1,...,\tilde{\F}^{(i)}_{\#\F}, \qquad i=1,...,K\] 
Evaluation of runs involves a new training/testing split of the dataset, and therefore the sequences $S_1,...,S_K$ generally will have feature variety. The set of feature sequences, $S= \{S_1, ...., S_K\}$ can then be given an index, 
\[ \mathcal{I}_S(k) = \frac{2}{K(K-1)}\sum_{1\leq i<j \leq K}\mathcal{I}_C(S_i(k), S_j(k))   \]
where, in the above, $S_i(k)$ denotes the first $k$ features selected by the $i$'th step of forward selection and $\mathcal{I}_C(S_i(k), S_j(k))$ is a measure of consistency between two feature subsets\footnote{Specifically, we have  $\mathcal{I}_C(\tilde{\F}_1, \tilde{\F}_2) = \frac{\# \tilde{\F}_1\cap\tilde{\F}_2\times \# \F - \#\tilde{\F}_1\#\tilde{\F}_2 }{\#\tilde{\F}_1(\F-\#\tilde{\F}_2) }$ for equal cardinality feature subsets $\tilde{\F}_1, \tilde{\F}_2$ }. With a stability index for the set of feature sequences, a final reduced feature subset can then be obtained by selecting an optimal subsequence and truncating its size based on the lowest cost (see \cite{stability} for full details).  The main advantage of this approach is that the returned feature subset should be fairly robust with respect to data samples.  
A fuller discussion of stability, with differing notions of measuring it, is found in \cite{stability}, to which we refer the interested reader.  We remark that, to the author's knowledge, no current open source automated feature engineering tools implement the stability-based approaches discussed here.  We include them as they may make a promising addition to future systems.

\section{Automated model engineering}
\label{model_engineering}

Model engineering  (as described earlier, a subset of autoML) can be thought of as consisting of at least two stages.  The first is the choice of a given model, $m$, from a candidate class of models, $\M$.  The second is the appropriate selection of hyperparameters $\pmb{\lambda}$, chosen from a set of hyperparameters $\Lambda$.  As outlined in  \cite{autoweka, autoweka2}, the hyperparameter space $\Lambda$ is necessarily hierarchical and often conditional; choices of a given hyperparameter $\lambda_j \in \pmb{\lambda}$ may highly influence other hyperparameters, $\lambda_i, i\neq j$.  For example, architectural modifications (limitations of depth or breadth in a neural network) may render other hyperparameters in changed network regions meaningless.  In fact, the second of these stages may be seen as a superset of the first; selection of which model to use may be cast as a choice of hyperparameter specification, though we may wish to keep them conceptually distinct.   Similar remarks of course hold for any preprocessing steps that are included in the pipeline; see \cite{hyperopt-sklearn}.  

A common trend in automated model engineering packages is the {\it pipeline}: an object joining multiple, conceptually independent processing steps into a cohesive unit able to be independently trained.  These pipeline components sometimes include aspects of model engineering and feature engineering as well as more banal data cleaning or imputation tasks.

The ultimate goal of automated model engineering is to eliminate the necessity of human involvement in the design stage of the model engineering process.  Users should be expected only to have to make very high-level choices, despite the complexity of the search space involved (both model and hyperparameter space can be effectively infinite).  In the below, we present some of the common concepts, algorithms, and open source Python package implementations used for automating the model engineering work of data scientists. For a fuller, more detailed history, the recent \cite{benchmark_survey}, \cite{automl}, and \cite{hpo} are excellent starting points covering many of the packages considered below. We refer the reader to Table \ref{model_table} for a comparison of models and methods of the automated model engineering packages we discuss in the following sections.

\subsection{Model selection and optimization}

In this section, we will be looking at the first-order issues involved with automatic hyperparameter and/or model selection, paying attention to some of the challenges unique to the overall model engineering problem and introduce some popular frameworks for solving them. Many of the packages discussed rely on techniques from Bayesian optimization and we refer the interested reader to Appendix \ref{bayesian_appendix} for explanation of some of the main concepts. 

\subsubsection{Ensemble selection} 
\label{ensemble_selection}
Perhaps the first automated system for tuning hyperparamters and selecting a model appeared in the form of {\it ensemble selection} introduced in \cite{ensemble_selection}. This technique consists of the generation of a large number of candidate models (originally, around 2000), obtained by variation of hyperparameters, without regard to performance quality. In an iterative and greedy, way models are evaluated to be added to an ensemble set based on how well the proposed model performs, when averaged with the existing ensembles, on a validation set.  In order to allow for continued performance gains, models are selected from the model population with replacement.  As well, ensembles can be initialized with a fixed number of top performers to avoid potential overfit.   Finally, the authors also explore {\it ensemble bagging}; constructing ensembles from a smaller random subset of the models to avoid overfit, repeating this process and finally averaging over all bagged ensembles. 

\subsubsection{Genetic techniques}
\label{genetic_tpot}
In \cite{biomedical_tree_based_pipwlinw}, the authors introduce {\bf Tree-Based Pipeline Optimization} (TPOT, see \cite{tpot_repo}), an algorithm that may have been the first to implement a full automated pipeline engineering solution that included feature transformation and selection.  The authors used the \textsf{DEAP} package to implement a genetic programming (see section \ref{feature_engineering}) algorithm generating trees of pipeline operators, which form the genetic population at any given stage of evolution.  The algorithm evolves the sequence of operators as well as the hyperparameters associated with each one and uses performance score on the learning task as a fitness function.  


The details of the basic \tpot algorithm allow for multiple copies of the data set to be propagated through a given pipeline before finally being used by an end stage classifier. As well, longer pipelines can be achieved than with packages like \textsf{auto-sklearn} which prohibit multiple preprocessing and model stages.  In the article \cite{evaluation_tree_based_pipeline}, the authors extend the basic \tpot framework, using Pareto optimization to regularize their model, penalizing needlessly lengthy pipelines. In \cite{scaling_tree_based} the addition of a feature subset selector and optional specification of linear or path-graph pipelines is explored, further enhancing \textsf{TPOT}'s abilities.

\subsubsection{Combined algorithm selection and hyperparameter optimization}
\label{cash_section}
In this section we review packages that implement solutions to a common formulation of the automated model engineering problem. 
\paragraph{auto-WEKA.}
In \cite{autoweka}, the authors introduce a {\bf combined algorithm selection and hyperparameter optimization} (CASH) scheme to find

\begin{equation}
\label{CASH}
 \argmin_{m_j \in \M, \pmb{\lambda} \in \Lambda^{(j)}} \frac{1}{k}\sum_{i=1}^k  \mathcal{G}(m^{\pmb{\lambda}}_j, D_{train}^{(i)}, D_{validation}^{(i)}) 
 \end{equation}
 
\noindent where  $\mathcal{G}(m_j^{\pmb{\lambda}}, D_{train}^{(i)}, D_{validation}^{(i)})$ denotes the incurred loss when training model $m^{\pmb{\lambda}}_j$, having parameters $\pmb{\lambda}$ that can take on values in a specified range $\Lambda^{(j)}$, on training subset $D_{train}^{(i)} \subset D\setminus D_{validation}^{(i)}$ and evaluating loss on validation set $D_{validation}^{(i)} \subset D$.  

The authors' implement a {\bf Sequential Model-Based Optimization} (SMBO, see \cite{scikit_optimize} for open source solutions) algorithm called Sequential Model-Based Algorithm Configuration (SMAC, see Appendix \ref{smac} for more details) to solve \eqref{CASH}.  SMBO is a general Bayesian technique for solving \textit{algorithm configuration} problems (see \cite{smbo}), attempting to maximize the expected loss improvement at a given hyperparameter choice with respect to a model-specific conditional probability.  Their implementation of this resulted in the \textsf{auto-WEKA} Java application which has Python implementations (see, e.g. \cite{pyautoweka}).  Since the original release of \textsf{auto-WEKA}, \textsf{auto-WEKA 2.0} (see \cite{auto-weka_2}) introduced several improvements on the original package, including support for regression models and a higher level of parallelism. In Table \ref{model_table} we refer to the various iterations of \textsf{auto-WEKA}, as the \textsf{auto-WEKA} family .

\paragraph{auto-sklearn.}
The \textsf{auto-sklearn} package \cite{auto_sklearn, auto_sklearn2} builds on the CASH formulation of \textsf{auto-WEKA} but incorporates two enhancements: {\it meta-learning} and {\it ensembling}. The meta-learning entailed first establishing meta-features and model performance on 140 datasets in the repository OpenML \cite{openml}, and using this information to assign a new dataset $D$ a similarity score to the known datasets, and using the associated model metadata for top performing models on the nearest 25 datasets to seed the Bayesian optimization portion of the CASH solver.  Evaluated meta-features are numerous and include things like numbers of instances/classes, ratio of categorical to numeric columns, some simple descriptive statistics, prediction performance with simple models, and others (see \cite{autosklearn_repo}).

The ensembling portion of the \textsf{auto-sklearn} algorithm keeps track of models during the Bayesian CASH optimization and then applies ensemble selection as described above.  The \textsf{auto-sklearn} package is capable of performing end-to-end modelling, including various preprocessing (imputation, balancing, etc) routines. 

In the article \cite{auto_sklearn2} the authors extend the original \textsf{auto-sklearn} package in a few ways.   On the prosaic end, they expand on datasets used for the meta-learning portion.  On the more challenging side, they consider a time-constrained version of the cash optimization \eqref{CASH}.  The main novelty used to satisfy the given time-constraint is to no longer use the meta-datasets to make a $K$-nearest dataset (KND) score on each new dataset given at runtime, but to use meta-datasets to construct a {\it pipeline portfolio} of pipeline configurations (hierarchical hyperparameters) that perform well in general, in terms of generalization error on validation subsets of the meta-dataset.  The portfolio obtained serves as a robust warmstart set for the Bayesian optimization used for solving CASH on a new dataset.  The authors introduced a {\it successive halving} technique for reducing the portfolio optimization time: train pipelines for a low given budget, select the best performing pipelines, increase their time budget in proportion to the number selected, and repeat.  This technique offers significant speedup on the time-constrained portfolio optimization.  The described variant of \textsf{auto-sklearn} is called \textsf{PoSH Auto-sklearn}. We refer to  \textsf{auto-sklearn} and its various variants as the  \textsf{auto-sklearn} family in Table \ref{model_table}.

\subsubsection{Other methods}
Finally, we present a few patterns for automated model engineering that do not fit neatly into any of the previous categories. 

\paragraph{FLAML.} In \cite{flaml} the authors introduce the \textsf{FLAML} (Fast and Lightweight AutoML) library (see \cite{flaml_repo, flaml_api}) which implements a few novelties in the algorithm selection and hyperparameter optimization space.  In \textsf{FLAML} a learning configuration is specified by a tuple $(m_j^\blambda, \blambda, r, s)$ where, as earlier, $m^\blambda_j$ is a machine learning model as usual, $\blambda \in \Lambda^{(j)}$, the associated hyperparameters for that model, $r$ is a binary variable indicating whether to use cross-validation or a holdout set (effectively determining how much of the training data is used in assessing performance) known as resampling, and $s$ is the sample size to use during training. The main novelty here is the decoupling of the sample size, resampling, hyperparameters and choice of model from each other allowing the potential exploitation of several heuristics\footnote{Some of these heuristics are the relationships between the resampling strategy and computation time, the relationships between smaller samples and need for regularization, etc.  See \cite{flaml} for the full discussion of observations and properties.}.

The method works according to the following.  First a resampling strategy is decided based upon the size of the training data, with cross-validation preferred for smaller sized datasets. After this, a machine learning model $m_j^\blambda$ is selected.  This selection is probabilistic with probability density determined based on the expected cost of improvement, a measure of the expected computation time to find a configuration of the given model with a lower validation error than the current lowest validation error among all models used thus far.
Models with a lower expected cost of improvement are expected to spend less time improving the overall performance and therefore are given a higher probability of being proposed. After the model is proposed, hyperparameters are selected using an algorithm described in \cite{cost} which uses a randomized stepping updates to $\blambda$ based on observed validation error. For a given model $m^\blambda_j$, starting with an initially small sample size, decisions are made about whether to increase the size with given choice of hyperparameters, or move to new hyperparameters with the current sample size. 

Although \textsf{FLAML} does not offer built-in ensembling as many of its competitors do, it manages to achieve solid performance.  Additionally it is highly extensible, offering users the ability to easily add in models, requiring very simple compatibility requirements. 

\paragraph{hyperopt-sklearn.} In \cite{hyperopt} the authors introduce the \textsf{hyperopt} package \cite{hyperopt_repo} for optimization in a variety of algorithm configuration problems.  Building on this framework, the \textsf{hyperopt-sklearn} package \cite{hyperopt_docs} uses \textsf{hyperopt} to performing pipeline selection from preprocessing and model modules implemented in \textsf{scikit-learn}.  The \textsf{hyperopt-sklearn} package offers users a high level of control over the method used for hyperparameter optimization, supporting a number of popular optimization algorithms such as random search (see, e.g., \cite{bergstra_search}) and annealing. The package is unique in allowing users a bit more control over the probabilistic aspects of the final estimators (probability of certain models being used, as well as distributions of hyperparameters).  

\paragraph{MLBox.} The \mlbox package \cite{mlbox_repo, mlbox_api} provides a fully automated, low-code wrapper for \hyperopt and \sklearn that includes multiple forms of feature selection (feature importances, LASSO, and others) as well as performing basic ensemble stacking.  

\paragraph{PyCaret}
The \pycaret package (see \cite{pycaret}) provides a very low-code, automated wrapper for \textsf{scikit-learn}, \textsf{hyperopt}, and others.  Heavily targeted at business case focussed data scientists, it offers some unique additions over competing packages, such as dedicated anomaly detection, classification, and natural language processing modules, robust preprocessing, as well as very good plotting and visualization tools with very minimal human input.  

\paragraph{OptimalFlow.} In \cite{optimalflow} the \textsf{OptimalFlow} automated pipeline ensembling package is introduced (see also \cite{optimalflow_docs}).  Unlike some of the other packages discussed so far, \textsf{OptimalFlow} used random search (\cite{bergstra_search}) or grid search rather than Bayesian optimization to solve the CASH problem.  \textsf{OptimalFlow} constructs a collection of pipelines to optimize, each with modularizable components, and then searches through the collection to obtain an optimal baseline in a technique the authors call Pipeline Cluster Traversal Experiments. 

%
%
%
%

%
%

\begin{table}[htp]
\centering
\resizebox{.85\textwidth}{!}{

    \begin{tabular}{  l p{10cm}  p{4cm}  p{5cm} }
        \toprule
\textbf{Package}      
& \textbf{Models}   
& \textbf{Ensembling}
& \textbf{Optimization} \\\midrule

\textsf{MLBox}&LightGBM, random forest, extra-trees, AdaBoost, linear&Stacking&Tree Parzen estimators\\

\textsf{OptimalFlow}&Logistic regression, SVC/SVR and linear SVC/SVR, linear, LassoCV, and ridge regression, SGDRegression/Classification, Huber, random forest, AdaBoost, kNN, MLP, XGBoost, HistGradientBoostingRegressor&No&Pipeline cluster traversal experiments, Grid search, random search\\

\textsf{hyperopt-sklearn}&Multiple SVCs/SVRs, kNN, AdaBoost, random forest, decision tree, multinomial and Gaussian naive Bayes, linear and quadratic discriminant analysis, one-v-rest/one, SGD, gradient boosting. &No & hyperopt (annealing, tree Parzen estimators)\\

\textsf{AutoGluon-Tabular}&LightGBM, CatBoost, XGBoost, random forest, extra-trees, neural networks, Vowpal Wabbit, DeepAR, Prophet, MQCNNEstimator, MLP time series estimator,  untested extensibility of full GluonTS models.& Multi-layer stacking& Fixed defaults, ensembling is optimization\\

\textsf{AlphaD3M}&AdaBoost, XGBoost, gradient boosting, random forest, logistic and least angle regression, nearest centroid, SVC and linear SVC, passive aggressive, kernel ridge, extra-trees&No&Monte Carlo tree search\\

\textsf{Prophet}&Additive regression Prophet model&No&Bayesian MAP estimation\\

\textsf{PyCaret}& Logistic and ridge regression, KNN, naive Bayes, linear and radial SVM, Gaussian process classifier, MLP, LightGBM, random forest, AdaBoost, linear and quadratic discriminant analysis, extra-trees, extreme gradient boosting, CatBoost&Stacking&Random search, Bayesian optimization\\

\textsf{EvalML}&KNN, random forest, XGBoost, LightGBM, CatBoost, extra-trees, decision trees, exponential smoothing, ARIMA, Prophet, Vowpal Wabbit&Stacking&Grid search, random search, Bayesian optimization\\

\textsf{OBOE}&LASSO and ridge regression, elastic net, KNN, decision tree, random forest, gradient boosting trees, AdaBoost tree, SVM, perceptron, Gaussian naive Bayes, extra-trees&Ensemble selection&Experiment design and collaborative filtering\\

\textsf{TPOT}&Gaussian, multinomial and Bernoulli naive Bayes,extra-trees, decision tree, kNN, XGBoost, logistic regression, SGD regression/classification, ElasticNetCV, AdaBoost, LASSO least angle regression, ridgeCV, MLP&Implicit stacking from synthetic features&Genetic\\

\textsf{AdaNet}&Neural networks&Complexity-regularized subnetwork ensemble&Boosting-style coordinate descent\\

\textsf{FLAML}&Random forest, extra-trees, LightGBM, XGBoost, ARIMA, SARIMAX, Prophet&No&Probabilistic expected cost of improvement and randomized direct search\\

\textsf{AutoTS}&Multiple naive models (zeroes, average value, seasonal, last value), GLS, GLM, exponential smoothing, UCM, ARIMA, VARMAX, DynamicFactor, DynamicFactorMQ, VECM, VAR, Theta, ARDL, Prophet, GluonTS, RollingRegression, WindowRegression, MultivariateRegression, UnivariateRegression, Univariate/MultivariateMotif, SectionalMotif, NVAR, NeuralProphet, GreyKite, MotifSimulation, TensorflowSTS, TFPRegression, ComponentAnalysis& Simple, distance, horizontal, mosaic& Genetic\\

\textsf{Auto-Keras}&Neural networks&No& Bayesian optimization with neural morphisms\\

\textsf{Auto-WEKA} family&BayesNet, DecisionStump, DecisionTable, GaussianProcesses, IBk, J48, JRip, Star, LinearRegression, LMT, Logistic, M5P, M5Rules, MLP, naive and multinomial naive Bayes,  OneR, RandomSubspace,  PART, random forest, RandomTree,  REPTree, SGD, SMO \& SMOreg, VotedPerceptron, ZeroR, LWL, AdaBoostM1, AdditiveRegression, AttributeSelectedClassifier,  RandomCommittee,&Stacking, voting&SMAC, Tree Parzen estimators.  Also supports grid and random search\\

\textsf{auto-sklearn} family &Adaboost, decision tree, extremely random trees, Bernoulli, multinomial and Gaussian naive Bayes, gradient boosting, kNN, linear and quadratic discriminant analysis, linear and kernel SVM, passive aggressive, random forest, SGD linear classifier &  Ensemble selection during search&Bayesian optimization warm-started with meta-learning \\

        \bottomrule
    \end{tabular}}
    
   \caption[Automated model engineering package comparison.]{Automated model engineering packages discussed in this survey. Some models have been omitted, though most were included.  Bayesian optimization, SMAC, and Tree Parzen estimators are discussed in Appendix \ref{bayesian_appendix}.  Included as well are time series based models, when appropriate.}
    \label{model_table}
\end{table}

\subsection{Second-order methods}
In this section we discuss extensions and enhancements of the CASH formulation described in Section \ref{cash_section}. These ideas in some cases touch on ``second-order" issues that arise in automating model engineering, touching at issues at a layer of abstraction higher than the basic CASH optimization \eqref{CASH} problem.  For a simple example of ``higher order" complications, while hyperparameters appear in machine learning models and require optimization, many AutoML systems address hyper-hyperparameter optimization (see, e.g. \cite{oboe} for this parameter hierarchy).  Each of the below packages and algorithms touch on higher-order abstractions in the automated model engineering problem.

\paragraph{Auto-sklearn 2.0.} We begin by remarking that the \textsf{PoSH Auto-sklearn} algorithm and package described above introduced additional meta-parameters (for instance, whether to run successive-halving in the optimization or not, and if so which parameters to set). The authors thus further extend \textsf{PoSH Auto-sklearn} by introducing a learned {\it policy selector}, which we now briefly describe (our notation will differ somewhat from \cite{auto_sklearn2}). 

The estimated generalization error of a model $m^\blambda$, on the meta-learning datasets $D_{meta}=\cup_j \{\tilde{D}_j\}$, is denoted $\widehat{Err}(m^\blambda, D_{meta})$ and is given by
\[\frac{1}{\# D_{meta}}\sum_{j} \widehat{Err}(m^\blambda, \tilde{D}_j) \]
where the error $\widehat{Err}(m^\blambda, \tilde{D}_j)$ is estimated on a holdout set in the usual way.  If $\pi$ denotes an optimization policy choice, from an allowable set of such choices $\Pi$, then the generalization error includes a dependence on the policy decision function $\xi: D_{meta} \to \Pi$ viz, 

\begin{equation}
\label{policy}
\widehat{Err}(m^\blambda, \xi, D_{meta}) \doteq \frac{1}{\# D_{meta}}\sum_{j\leq \# D_{meta}} \widehat{Err}(m^\blambda(\xi(\tilde{D}_j)), \tilde{D}_j)
\end{equation}
with $m^\blambda(\xi(\tilde{D}_j))$ denoting the model $m^\blambda$ with optimization policy on $\tilde{D}_j$ determined by $\xi$.  The second order learning process then involves optimizing over the policy space in addition to the model space.  Cross-validation {\it of the meta-data sets themselves} is performed, building each policy on subsets of $D_{meta}$ and validating on a holdout subset, in order that the estimated performance of a policy on any given dataset be reliable.  The authors of \cite{auto_sklearn2} use a combination of techniques to then determine an appropriate $\xi$.  The resulting open source implementation of this enhancement to \textsf{Auto-sklearn} is the \textsf{Auto-sklearn 2.0} package, available as a module within the   \textsf{Auto-sklearn} package \cite{autosklearn_repo}.

\paragraph{AutoGluon-Tabular.}
In the article \cite{auto_gluon} the authors introduce the \textsf{AutoGluon-Tabular} fully automated model engineering algorithm and package (see \cite{auto_gluon_repo}) capable of extremely low-code generation of machine learning models.  Data is passed through both model-agnostic and model-dependent preprocessing stages prior to being input to a number of sequentially trained models chosen from a broad class, including Extremely Randomized Trees and neural networks.  The model training is scheduled to allow good results to obtain even under heavy time-constraints.

%
%
%

An important part of \textsf{AutoGluon-Tabular} is the use of a {\it multi-layer stacking} technique for model ensembling (see also \cite{stacked_generalization}). While packages like \textsf{EvalML} and \mlbox use a common stacking ensemble technique consisting of training a final ``stacker" model, whose inputs are the predictions of the ``base" learners in the candidate model space, on a holdout set, in  \textsf{AutoGluon-Tabular} stacker models are layered, with outputs of a given set of stacker models serving as inputs to the next layer of stackers.  The higher-level stackers are trained, for a given base model, only on the holdout predictions, never on in-fold predictions, as data is partitioned into $k$ disjoint chunks.  To reduce potential overfit, this $k$-fold bagging process is repeated a number of times and the average of the out-of-fold predictions is the final estimator at a given level of stacking.

%

\paragraph{AlphaD3M.}
In the article \cite{alphad3m}, a team from New York University introduced \textsf{AlphaD3M}, a pipeline selection and hyperparameter optimization package \cite{alphad3m_repo} built around DARPAs Data Driven Discovery of Models (D3M) program \cite{d3m}.  As apparent in the name, \alphad takes heavy algorithmic inspiration from \textsf{AlphaZero}, a program developed by DeepMind that masters several two-player games by learning through self-play.  Training examples for \alphad  are created by the system during self-play,  and model pipeline primitives (feature extraction, inference, etc) were taken as fundamental modular units in the process. A long short-term memory (LSTM) network is trained to predict action probabilities corresponding to a pipeline distribution as well as estimated performance. A regularized cross-entropy loss on the predicted pipeline sequence (again, viewed as drawn from a distribution) and a pipeline found from Monte Carlo Tree Search (MCTS) together with squared errors from expected and obtained performance is then used to optimize the network, separately penalizing cumbersome pipelines and network weights. 

In \cite{alphad3m2} the authors extend the basic \alphad framework discussed above by incorporating a context-free grammar over for valid pipeline construction, based on practitioner knowledge about pipelines.  With this pipeline grammar, the MCTS computation time is significantly decreased.  This improvement showed that even a small amount of expert knowledge can have profound gains for autoML systems.  The authors additionally showcased the impact that pre-trained models can make on \textsf{AlphaD3M}. 

\paragraph{EvalML.} The \evalml package (see \cite{evalml}) developed by Innovation Labs at Alteryx performs full model engineering and is designed to be highly compatible with \textsf{Compose} and \textsf{featuretools}. Aside from doing automated pipeline optimization, ensemble stacking, and excellent graphical pipeline visualization options, \evalml also offers {\it explainability} modules.  These allow users to explore local interpretable model-agnostic explanations (LIME, see \cite{lime}) and  SHAP values to gain insight into the decisive features in a selected prediction.  These techniques assess the relative impact of features in a {\it particular decision}, as opposed to the feature importances, which measure the relative importance of a feature in the overall model. 
This, together with the natural language explanations offered by \textsf{featuretools} make an end-to-end \textsf{featuretools}/\evalml pipeline an attractive choice when explainability and transparency are important considerations (e.g. use cases in highly regulated industries).

\paragraph{Collaborative filtering.}
In \cite{oboe} a method based on collaborative filtering (see \cite{funk} for a very good explanation of this class of problem) is proposed for solving the algorithm selection and hyperparameter configuration problem resulting in an algorithm and package called OBOE\footnote{The name derives from the woodwind instrument orchestras use to provide tuning.} (see \cite{oboe_repo}).  The main idea can be summarized as follows below. 

Let $\{\tilde{D}_i\}_{i \leq m}$ and $\{m_j\}_{j \leq n}$ be a collection of meta-datasets and models, respectively.  A matrix $M$ can be constructed such that 
\[ M_{ij} = \text{cross-validation error of } m_j \text{ using } \tilde{D}_i\]
The matrix $M$ can then be approximated by a low rank matrix comprised of latent meta-features, i.e. $M_{ij} \approx \mathbf{w}_i \cdot \mathbf{z}_j$  for $\mathbf{w}_i, \mathbf{z}_j \in \R^k$, $k <\min(m,n)$, latent feature vectors of meta-datasets and models, respectively, which can be computed, essentially, using singular value decomposition.  When provided a new, previously unseen, dataset $D_{new}$, some models are run on it and associated cross-validation errors are computed, effectively appending a new, partially filled, row to $M$.  This occupies the majority of the online runtime in \textsf{OBOE}. Latent meta-features for $D_{new}$ are then obtained via
\[\mathbf{w}^{(new)} = \argmin_{\mathbf{w} \in \R^k}\sum_j (e_j - \mathbf{w}\cdot \mathbf{z}_j)^2 \]
where the sum is taken over the $j$'s corresponding to models for which errors were calculated for $D_{new}$.  The performance of (unobserved) model $l$ on $D_{new}$ can then be predicted using $\mathbf{w}^{(new)}\cdot \mathbf{z}_l$. Additionally, the authors regress model runtimes on sample size and features in each of their meta-datasets allowing for surprisingly accurate estimation of unobserved model runtimes.  Notice that the method used by \textsf{OBOE} assumes a bilinear relationship between model performance and meta-features and models.

The only piece missing from above is how the models to run on a new dataset are chosen.  For this, the authors use a $D$-optimal (see, e.g. \cite{d_optimal} for related problems) method inspired by experiment design, in which a constrained (with total estimated runtime constraints) convex optimization problem is solved to produce the subset of models to use in order to estimate $\mathbf{w}^{(new)}$. Once the model performances on $D_{new}$ have been either observed or predicted, an ensemble selection process (see \ref{ensemble_selection}) builds an ensemble learner on the top observed performers\footnote{A priority queue is used, based on predicted performance, to decide which models to actually observe.}.  

While \textsf{OBOE} models the model/feature to performance relationship linearly, in \cite{pmf}, the authors instead allow for a nonlinear relationship by probabilistic matrix factorization (PMF, see \cite{nonlinear_pmf}).  For this, the authors look to model $M$ using an explicitly nonlinear relationship, $M_{ij} = g_j(\mathbf{w}_i) + \epsilon$, where $\epsilon$ is a Gaussian distributed random variable with variance $\sigma^2$ and $g_j$ is a nonlinear function modelled using a Gaussian process prior (see Appendix \ref{gaussian_processes}).  In this case \[(g_1, ...., g_n)\mid \begin{bmatrix}\mathbf{w}_1\\\vdots\\\mathbf{w}_m \end{bmatrix} \sim  \normal( 0, K) \]
for covariance matrix $K$'s entries given by kernel function $k(\mathbf{w}_i, \mathbf{w}_j) = K_{ij}$ and $\normal$ a multivariate Gaussian distribution.  Putting all parameters associated with kernel function $k$ in $\pmb{\theta}$ gives the following likelihood function for the performance matrix $M$,
\begin{equation}
\label{pmf}
L(W, \pmb{\theta}, \sigma^2) \doteq \prob(M\mid W, \pmb{\theta}, \sigma^2) = \Pi_{i \leq n} \normal(M_{:, i} \mid 0, K + \sigma^2 I)
\end{equation}
where $M_{:,i}$ denotes the $i$'th column of $M$ and $W$ is a matrix made up of the latent features $\{\mathbf{w}_i\}_i$ as columns. The likelihood function in \eqref{pmf} can be optimized using stochastic gradient descent to reveal the latent features in $W$.  In this framework, predictions according to new models will also follow a normal distribution and depend on previously evaluated models.  The selection of subsequent models to evaluate is done via maximizing the expected (performance) improvement acquisition function.  An implementation of this technique is found in \cite{pmf_repo}.  Despite the seeming increase in expressiveness provided by PMF described here, comparisons between \textsf{OBOE} and PMF (see \cite{oboe}) indicate that the low-rank, linear assumptions of \textsf{OBOE} provide solid competition against PMF based model selection.

\subsection{Automated forecasting}
As with feature engineering, time series present their own unique challenges when it comes to automating the model engineering process.  We can only briefly touch on some of the additional complications involved with automated forecasting; for a recent, thorough, review of some of the challenges and solutions in this space, we refer the reader to \cite{time_series_review}.  The unique challenges with time series forecasting we wish to mention arise from a number of sources including the following:
\begin{itemize}
\item {\bf Model Variety.} A large number of available models is not an issue unique to time series, however time series have their own classes of models designed specifically for forecasting purposes.  These are generally naive models which rely on simple techniques, statistical methods which include autoregressive and smoothing models, and machine learning models (see \cite{deep_learning_time_series}) which train regression models using more traditional machine learning techniques not directly inspired from classical time series modelling. 
\item {\bf Assumptions.} Classical time series models generally have their own types of assumptions which may include different forms of stationarity of the series as well as ergodicity (see, e.g. \cite{brockwell}). This means that automated forecasting tools should be able to detect and, potentially correct (as part of an overall pipeline) non-stationarity.  
\item {\bf Forecasting.} Time series forecasting generally requires more than single value prediction and, rather, usually involves prediction of a sequence of values into a future horizon. Forecasting models therefore must be able to reliably capture more than just simple responses and must also capture the {\it order-dependency} of the predicted values. In addition, it may be desirable to forecast probabilistically by providing either forward-looking confidence bounds or future distributions. 
\end{itemize}
Each of the issues mentioned above can be addressed by their own selection of different methods. For instance, there are several techniques available for detecting and addressing nonstationarity in time series\footnote{The article \cite{time_series_review} gives a nice overview of many of the techniques available for this. We just remark that ergodicity is, in general, not something easily verified.}.  Each available option of technique for the above issues can represent a pipeline component, potentially carrying their own hyperparameters just as we saw in non-time series prediction problems\footnote{For instance the choice of $p$ and $q$ in an $ARMA(p,q)$ model, the backwards horizon period over which to train a machine learning model, or choice of distributional parameters in a Bayesian model to name just a few examples. }.

\paragraph{Prophet.} Inspired by generalized additive models (see \cite{gam}), in \cite{prophet}, the authors introduce the \textsf{Prophet} model for automated time series forecasting (see \cite{prophet_repo}) attempting to provide a scaling of the tools in this space in the sense of expanded userbase, variety of forecasting problems, and efficient evaluation of large numbers of forecasts.  The model they introduce, 
\begin{equation}
\label{prophet}
y(t) = g(t) + s(t) + h(t) + \epsilon_t
\end{equation}
is made up of a nonlinear growth term, $g$, a seasonality term, $s$, holidays, $h$, and a Gaussian noise term, $\epsilon_t$.  The growth term the authors use is a modified logistic model with nonconstant carrying capacity and piecewise constant growth rates\footnote{Their framework also allows for unbounded growth models.}, specified by rate changepoints that use Laplace-distributed priors.  The seasonality term in \eqref{prophet} is modelled on a truncated Fourier series, and takes the form $s(t) = X(t)\pmb{\beta}$ with $X(t)$ a vector of Fourier basis elements and $\pmb{\beta} \sim \normal(\mathbf{0}, \sigma^2 I)$. Holidays are treated as independent shock events and models as $h(t) = (\mathbbm{1}_{t \in holiday_1}(t), ...., \mathbbm{1}_{t\in holiday_k}(t))\pmb{\kappa}$ for $\kappa \sim \normal(\mathbf{0}, \nu^2 I)$ where $holiday_j$ is a set of dates for the $j$'th holiday. With these priors on each component in the additive model \eqref{prophet} a maximal a posteriori estimate can be used to determine parameters and forecast or a generative posterior can be used to forecast with uncertainty. The implementation of the \textsf{Prophet} model \cite{prophet_repo} is highly low-code but also allows interested users control over lower-level parameters.

\paragraph{AutoTS.} The \textsf{AutoTS} package (see \cite{autots}) is a robust, genetic algorithm based automated forecasting library that focuses on providing highly low-code time series models.  The model families used by \textsf{AutoTS} are large and include naive, statistical and deep learning members. The package allows for a variety of validation techniques as well as the ability to handle multiple time series, probabilistic forecasts, extensive control over the scoring of models via weighted combinations of metrics\footnote{As the authors of the package note, this composite scoring technique (which works well in the genetic algorithm paradigm since differentiation is not needed) allows for \textsf{AutoTS} to balance competing, potentially exclusive performance criteria in an elegant way. See the extended example in \cite{autots} for a fuller discussion.}, and control over the overall speed of learning by restricting the allowable model types.  In addition, \textsf{AutoTS} offers several forms of ensembling: {\it simple} ensembling is a basic averaging approach, {\it distance} is the splicing of the forecasted data into separate forecast regions, one for each model projection, {\it horizontal} in which models and parameters are independent across the multiple time series, and {\it mosaic}, which extends {\it horizontal} ensembling by having each forecast period for each series get its own model.  Since the {horizontal} ensembles involve lower level models, they can involve recursively {simple} or {distance} ensembled forecasts.

\paragraph{GluonTS.}
In the article \cite{gluonts} (see also \cite{gluonts_jmlr}), the authors introduce the time series modelling package \textsf{GluonTS} \cite{gluonts_repo}, which implements a host of probabilistic and deep learning models built in \textsf{MXNet} and \textsf{PyTorch}, among others.  The models in \textsf{GluonTS} are, in the main, divided into generative and discriminative types\footnote{The particular meaning of these terms in \cite{gluonts} differs somewhat from the usually encountered meanings, as in \cite{ng_jordan}.} with Deep State Space Models (see \cite{deep_state_space}) being one example of the former and Transformers (see \cite{transformer}) an example of the latter. 
The \textsf{AutoGluon} library has a forecasting module incorporating several deep learning models from \textsf{GluonTS} including convolutional encoder-decoder network (see \cite{mqcnn} or \cite{deep_learning}) and potential (untested) ability to add any \textsf{GluonTS} estimator object into the \textsf{AutoGluon} framework.

We close this subsection on time series by noting that, in addition to the above mentioned packages, \textsf{PyCaret}, \textsf{EvalML}, \textsf{FLAML} and \textsf{Auto-Keras} (see next section) offer the users ability to do forecasting as part of their general model engineering suite.  The models offered within these packages are outlined in Table \ref{model_table}.

\subsection{Deep AutoML}
The automated model engineering packages considered thus far have concentrated on more classical models (decision trees, linear regression, support vector machines, etc) as opposed to neural networks.  Because of the complexity of even basic feedforward neural networks\footnote{In a standard fully-connected feedforward network with architecture defined by layers $L_0, ...., L_f$, with $L_0$ being the input layer and $L_f$ being the output layer size, and corresponding depths  $d_0, ..., d_f$ has $\sum_{i=1}^f (d_i d_{i-1} + d_i)$ parameters, excluding any hyperparameters associated with initialization of regularization.  For common input vector sizes and just a single hidden layer, this can be quite a lot of parameters.}, automatic generation of a network for a specific task presents its own set of unique algorithmic challenges, not least of which is the increased time costs associated with training candidate models. In the following we cover a few open source packages and their underlying algorithms for tackling this complex problem which is sometimes known as AutoDL (e.g. \cite{auto_pytorch_1}). While some of these packages present users with ``only" neural networks as candidate models, universal approximation theorems (e.g. \cite{cybenko}) ensure the profound robustness of the resulting space.  We refer the reader to \cite{deep_learning} for more thorough coverage of the complexities involved with neural networks more generally.

\paragraph{AutoKeras.}
 In the article \cite{autokeras} the authors introduce an algorithm for {\bf neural architecture search} (NAS, see \cite{nas}) to automatically optimize the architecture of a neural network as well as introduce the \autokeras package implementing this algorithm \cite{autokeras_api}. 

The particular approach to the NAS problem taken by \autokeras relies on neural morphisms (see, e.g. \cite{neural_morphisms}), transformations between networks that preserve functionality, as an essential part in the Bayesian optimization.  The search space in the optimization routine is effectively a tree whose edges are morphisms between nodes representing architectures.  In order to use Bayesian optimization, the authors circumvent the need for Euclidean representation for the underlying Gaussian Process and instead produce a kernel directly, obtained from an embedding of an approximation of the edit-distance between two proposed networks (see \cite{graph_edit}).  The acquisition function optimization is handled by a bespoke version of $A^*$ search (see \cite{ai}), returning a path through the tree to obtain an architecture (i.e. a sequence of morphisms to apply). An additional challenge was enforcing shape consistency constraints.

\paragraph{AdaNet}
In \cite{adanet} an algorithm for NAS is presented, whose implementation, \textsf{AdaNet} (see \cite{adanet_docs}) uses, as implied by its name, an adaptive technique for generating new architectures from previous ones.  The architecture space includes any neural network that can be represented as a directed acyclic graph (DAG), and thus includes skip-connections and more exotic layouts. 

The main idea behind the \textsf{AdaNet} algorithm is to iteratively modify the architecture in a way that optimizes a very specially designed objective.  To wit, an objective is used that consists of a convex surrogate loss with penalization on complexity (specifically, Rademacher complexity\footnote{Given a data sample $D = \{x_i = X(\omega_i)\}_i$, $g$ in a function class $\mathcal{G}$, and $\{\sigma_i\}$, $\{-1, +1\}-$valued identically distributed Rademacher variables, i.e. $\prob(\sigma_i=1) = \prob(\sigma_i = -1)$, for all $i$, the Rademacher complexity of the class $\mathcal{G}$ is defined by 
\[ \mathcal{R}(\mathcal{F}) = \frac{1}{\# D} \mathbb{E}\{\mathbb{E} \sup_{g \in \mathcal{G}}\sum_{i \leq \# D}\sigma_i g(X(\omega_i))\}\] 
where the expectations are over the Rademacher distribution and the data generating distribution.  The Rademacher complexity measures the ability of a function class to model noise as it is the most expected correlations with noise a given function class should be able to obtain.  In other words, it captures how robust the function class is to random labelling of data in a binary classification problem. See, e.g.,\cite{rademacher} for a fuller description.} 
).  At each iteration of the algorithm, candidate weak learners (potential subnetworks) are generated by a module, and the network is augmented by a candidate if it causes a reduction in the objective.   In this way, the final network evolves from the accumulation of multiple weak learners, with augmentations happening in a way that allows new candidates to learn from features generated by prior iterations.

\paragraph{Auto-PyTorch}
In \cite{auto_pytorch_2} the authors introduce a system, \textsf{Auto-Net}, instantiated in two packages, \textsf{Auto-Net 1.0}, a SMAC-based solution to \eqref{CASH} based on \autoweka and \textsf{Auto-Net 2.0}, or \textsf{Auto-PyTorch} (see \cite{auto_pytorch_repo}), which uses a combination of Bayesian optimization and the bandit-based HyperBand (HB, see \cite{hyperband, hyperband_2}) algorithm, known as BOHB \cite{bohb}.  \textsf{Auto-PyTorch} works with four different neural network types; multi-layer perceptrons with dropout \cite{dropout}, residual networks \cite{resnet}, in both shaped and nonshaped versions, with different learning schedulers and optimizers available.  

BOHB uses kernel density estimation to estimate opportunistic regions in the algorithmic conifiguration space (both architectural and associated hyperparameters) and has the advantage of being easily parallelizable. It works on so-called {\it multi-fidelity} optimization problems in which multiple budget (e.g. runtime or epochs) granularities can be explored with performance-based promotion to increased budget size. Smaller-budget configurations then help guide the search and using sampling from the kernel density estimator promotes variety in the resultant configurations. 

In a followup article, \cite{auto_pytorch_1}, a refinement and specialization of \textsf{Auto-PyTorch}, \textsf{Auto-PyTorch Tabular} is introduced, focussing specifically on structured data most relevant in traditional data science applications.  The resulting algorithm includes a similar ensembling strategy as used by \textsf{auto-sklearn}, as discussed in Section \ref{cash_section}, with the additional ability of being able to incorporate other classes of models into the ensembler. As well, warmstart for the BOHB is provided by configurations selected for performance on meta-training sets by a portfolio optimization strategy similar to that used in \textsf{auto-sklearn PoSH}.

\section{Remaining challenges}
\label{remaining}
While there has been tremendous progress made in recent years on increasing automation into the general data science workflow, there still remain challenges that are not, in our opinion, adequately solved by existing open source libraries.  We highlight a few areas where we feel progress can still be made that would result in significant gains for the data science community. 

\paragraph{Problem framing.} As noted in Section \ref{overview} the three automation steps (autoPE, autoFE, and autoME) presuppose a (precise) formulation of a given prediction problem.  Not all problems lend themselves to unambiguously specified formulations.  We consider here two very simple examples:
\begin{itemize}
\item First, we consider customer churn (see, e.g. \cite{churn}).  The time period over which a customer may be considered lost or retained will vary considerably on a case by case basis.  As well, segmentation of customers will often be an important step needed to have a reliably meaningful churn rate (see \cite{churn_2} for this and  a discussion other problems with simple churn rates).  Both the time period involved and the segmentation are, essentially, determined by human judgment and/or business logic and therefore not easily automated.  

\item Consider next the issue of predicting {\it promotions} within an organization.  In a large enough organization, it may not be a straightforward task to determine what constitutes a promotion.  Suppose the following holds: department $X$ has titles {\it Entry}, {\it Junior}, {\it Senior} and {\it Executive} while department $Y$ has the titles {\it Entry}, {\it Associate}, {\it Lead} and {\it Executive}.  All pay bands are higher (and non-overlapping) in department $Y$ than their counterparts in department $X$ due to geographic or product-focussed reasons\footnote{The human resources department may not make such distinctions and just assign the levels to numbers 1--4 or similar.  Whether they do this or not, both raise additional potential complexities.}.  Bob in department $X$ at {\it Junior} level is internally transferred (as a result of merit) to an {\it Associate} position in department $Y$ while Alice moves from a {\it Lead} role in department $Y$ to an {\it Executive} role in department $X$ (resulting in a pay decrease). Also, Carol left department $X$ in a {\it Senior} role in January only to be rehired as an {\it Executive} in $X$ in February.  The precise logic as to whether Alice, Bob, and Carol were or were not promoted does not seem separable from specific business knowledge and therefore not easily automated. 

\end{itemize}
As these basic examples illustrate, even very simply-stated prediction problems necessarily entail value judgments, business or domain knowledge, and sometimes arbitrary decisions to formulate in precise terms. These preclude immediate progress on automating this first step in the data science problem lifecycle.

\paragraph{Early stages.} As discussed in Section \ref{overview} there are stages close to the data ingestion stage where data scientists often spend significant time for which automated open source tools do not seem readily available. These include class balancing by sampling or synthetic generation (e.g. SMOTE \cite{smote}), more advanced methods of imputation of missing values, detection and/or removal of outliers, stratification of the observations, detection of drift.  Some of these tools may be automated and embedded in a larger ingestion or cleansing pipeline upstream of automated data science predictions, however this is often not the case and data scientists may spend significant time on several of these steps within a given project.   While some of the AutoML tools we've investigated address some of these issues (most handle simple univariate imputation and normalization, for instance) none seem to address the more sophisticated techniques in the data scientist's general arsenal for tackling these preliminary preprocessing tasks.

\paragraph{Scalability.} Machine learning models can form the primary analytical engine of data science prediction pipelines, however, they are still, in many cases, a small part of the overall system.  How well a machine learning model fits into the existing production infrastructure is a critical factor in determining adoption (see \cite{turin} for a nice summary of the issues involved).  This may involve some of the following considerations: latency, built-on data-checks and data monitoring, expected computation resources, uncomplicated compatibility with data ingestion protocols, ease of incorporation into A/B testing paradigms, flexibility of model types and objectives used, native parallelism and/or ability to work easily in a distributed compute environment, transparency and interpretability of predictions (see \cite{trust} for more on this topic).  Integrating more seamlessly into existing technological pipelines can be, depending on the context, as important as predictive performance. 

While the preceding issues are critical for any automated prediction engine, they are particularly important in the open source libraries considered in the article as there do exist enterprise-level AutoML platforms devoted entirely to helping address various of these concerns and thereby increasing the odds of AutoML making it into a deployed solution. While these closed source solutions are not part of our focus here, some common ones are \cite{amazon, datarobot, azure, ibm} and \cite{alteryx}.

\paragraph{Usability.} Many of the packages explored in the preceding are low-code (e.g. \textsf{PyCaret}, \textsf{MLBox}, and \evalml are particularly good on this front), however, most generally do require some technical expertise on the part of the user.  Installation is not always a straightforward task, depending on the platform and package, this being particularly true with \textsf{Auto-WEKA}, \textsf{h2o}, and \textsf{auto-sklearn}. As discussed in \cite{volcanoml}, many AutoML search spaces may not be adequate for many real-world applications.
 
Conversely, more advanced users may wish to have more fine-grained control over some aspects of the automation than are offered by the package. The AutoML packages discussed in the preceding offer users differing degrees of control over parameters that may users may wish to adjust or maintain.

\paragraph{Reusability and modularity.} Most of the AutoML systems discussed in this survey exhibit limitations in terms of their general reuse.  For example, expert or domain knowledge can be a crucial part of feature engineering and, as a consequence,  automating the feature generation and selection process may preclude pipeline reuse being effective for a wider variety of data science contexts.  An additional potential impediment for having more robust systems is the lack of modularity of specific pipeline components.  Isolating the feature selection stage from one automated system, say, might require running the entire AutoML process, which may be prohibitive in addition to unnecessary.

\paragraph{Rigidity.} Lastly, data scientists often work on a variety of different use cases. In some cases, they handle data in multiple forms, work on varying timelines, face a range of reporting requirements, and even program in different programming languages and  environments.  Many times, this variety takes place within a single project.  Because of this\footnote{And, surely, other unnamed reasons.}, data scientists often adopt idiosyncratic working styles, developed based on subjective criteria such as familiarity, adoption by a larger team, perceived learning curve and time investment to replicate existing functionality, lack of organizational investment in or commitment to promoting growth (see, e.g. \cite{hbr}),
and other forms of psychological inertia or rationalization\footnote{The rationalization may or may not be justified.  It may well be the case that performance, for instance, will be improved with more human involvement and less automation, on any given project. Whether the performance gains justify the extra costs is highly context-specific.}.  Done right, data science reinforces deep scientific skepticism (about new datasets, new trends, new models, new deadlines, new technology, etc) and can foster caution about new, untested adjustments to existing patterns.  Automation plays directly against these forces by removing the control and specificity provided by carefully designed or optimized workflows developed through hard-won experience and continuous iteration.

%
\section{Acknowledgments}
I am indebted to the helpful suggestions and improvements offered by D\'avid Par\'azs, Thiam Lee, and Hao Chang during their careful review of an earlier draft of this report.  Figure 1 was created by Tatiana Avaeva.

\bibliographystyle{plainurl}
\bibliography{ai_references}

\appendix
\section{Bayesian optimization primer}
\label{bayesian_appendix}
Because Bayesian optimization plays such a large role in several of the automated model engineering packages discussed in this article, we provide the interested reader with a self-contained, though superficial, review of the basics of the approach, together with the specific variants relevant to the automation process considered above.  We recommend \cite{practical_bayesian} and \cite{bayesian_tutorial} for further reading on Bayesian optimization more generally. 

\subsection{The black box optimization problem}

Bayesian optimization is concerned with the classical objective optimization problem
\[ \argmax_{\x \in \Omega} f(x)\]
where the objective function\footnote{As in equation \eqref{CASH} minimization problems can similarly be cast into this framework by using $-f$ in place of $f$.}, $f: \Omega \subset \R^n \to \R$.  Specifically, it was designed to handle cases were the functions fail to have properties expected of more traditional techniques; the objectives may be non-convex, difficult to find closed form expressions for, non-smooth, expensive to evaluate (as will be the case when the inputs are hyperparameters for a machine learning model that needs to be trained), or otherwise unconventional.  For this reason, Bayesian optimization is often referred to as a {\it black box} optimization technique since it does not require explicit (analytic) expressions of the objective. 

To deal with these complexities, in \cite{mockus_joint, mockus_sole} the authors describe a broad class of optimization problems based around viewing the objective function as stochastic; namely, $f(\x)$ may be viewed as shorthand for $f(\x,\omega)$ where $\omega$ is  in a  $\sigma$-algebra.  The function $f$ then carries with it an associated distribution function
\[ F_{\x^{(1)}... \x^{(m)}}(y_1, ..., y_m) \doteq \prob(f(\x^{(1)})\leq y_1, ..., f(\x^{(m)})\leq y_m)   \]
induced by observations $\{\x^{(i)} \}_{i=1}^m$.   Defining $\z_k = \cup_{i\leq k}\{\x^{(i)}, f(\x^{(i)}) \}$\footnote{Strictly speaking, the observations may be noisy, and therefore of the form $ \cup_{i\leq k}\{\x^{(i)}, \normal(f(\x^{(i)}), \xi)\}$ for a choice of $\xi \in \R$. } to be a set of $k$ observations of the function's input/output pairs, Bayes' formula becomes the following
\begin{equation}
\label{bayes}
\prob(f \mid \z_k ) = \frac{\prob(\z_k\mid f)\prob(f)}{\prob(\z_k)} \propto \prob(\z_k\mid f)\prob(f)
\end{equation}

The left-hand side of equation \eqref{bayes} represents the posterior distribution of the function $f$, given the $k$ observations in $\z_k$.  The prior distribution, $\prob(f)$, represents initial assumptions about the objective (smoothness, convexity, etc) and influences the likelihood function, $\prob(\z_k \mid f)$.

\subsection{Detour on Gaussian processes}
\label{gaussian_processes}
It is common in Bayesian optimization to use a {\bf Gaussian process} prior distribution for $f$.  A Gaussian process, $\{X_t\}_{t\in \mathcal{T}} = \{X_t(\omega)\}_{t\in \mathcal{T}}$ with $\mathcal{T}$ an (possibly multidimensional) index set, is a stochastic process with the property that any finite subset of the process is multivariate normally distributed (see, e.g. \cite{gaussian_processes}). As multivariate normal distributions are fully characterized by their mean vectors and covariance matrices, the same is true of a Gaussian process. Namely,  $\{X_t\}_{t\in \mathcal{T}}$ represents a Gaussian process if and only if $\{X_t\}_{t \in \mathcal{I}}$, with $\# \mathcal{I} < \infty$, satisfies that $(X_{\sigma(\mathcal{I}_1)}, ..., X_{\sigma(\mathcal{I}_{\# \mathcal{I}})}) \sim \normal(\pmb{\mu}, \pmb{\Sigma})$, for some $\pmb{\mu}$ and symmetric positive-definite $\pmb{\Sigma}$ and for all permutations $\sigma$ on sets with $\#\mathcal{I}$ elements.  
Viewing each such vector as ordered samples from a function, it is convenient to say that a Gaussian process $g: \R^n \to \R$ is a function for which any finite sample of its values represents draws from a multivariate normal distribution.  In that case, we have that 
\begin{align*}
\mu(\x) &= \expectation[g(\x)]\\
k(\x, \tilde{\x}) &= \expectation[(g(\x) - \mu(\x))(g(\tilde{\x})-\mu(\tilde{\x}))] = Cov(g(\x), g(\tilde{\x}))
\end{align*}
define the mean and {\it kernel}, $k$\footnote{The kernel function induces spatial correlation on $\R^n$ via $g$. },  of a Gaussian process, respectively, and write \[g\sim \mathcal{G}\mathcal{P}(\mu, k)\] 
A few remarks are in order before concluding this subsection: 

\begin{itemize}
\item If a Gaussian process is restricted to having a finite index set, it follows from the definition that the process is simply a multivariate Gaussian distribution.  
\item Fixing a single spatial point, $\x$, $g \sim \mathcal{GP}(\mu, k)$ means that $g(\x)$ is normally distributed.  Therefore, by varying $\x$, we see the assumption of $g \sim \mathcal{GP}(\mu, k)$ amounts to viewing the Gaussian process as supplying a ``random function", its expected magnitude determined by $\mu$, and its unpredictability determined by $k$.  
\item Often the kernel functions, $k$, will have hyperparameters associated with them which may be associated with correlation range or anisotropies.  Therefore, writing $k_{\pmb{\theta}}$ will make explicit the dependence of the kernel on hyperparameters $\pmb{\theta}$.
\item Gaussian processes are closed under sampling; if $f \sim \mathcal{GP}(0, k)$ then it can be shown that $f\mid \z_k \sim \mathcal{GP}(\mu, \tilde{k})$ with $\mu, \tilde{k}$ having definite analytic expressions (see \cite{bergstra_algos}).
\end{itemize}

\subsection{The Bayesian algorithm}

As mentioned, a common Bayesian optimization prior is then to assume that the objective function $f$ is an unknown Gaussian process, namely $f \sim \mathcal{G}\mathcal{P}(\mu, k)$ for given choice of mean and kernel function. Sampling $f$ helps  to remove uncertainties in the associated Gaussian Process model via Bayesian updates. 

In the following, we will assume $\mu = 0$ (a common simplifying assumption, see e.g. \cite{bergstra_algos}) and follow \cite{bayesian_tutorial} closely, By an abuse of notation, denote $\mathbf{f}_k = (f(\x^{(1)}), ..., f(\x^{(k)}))^T$, then $\mathbf{f}_k \sim \normal(0, K)$ with $K_{ij}=k(\x^{(i)}, \x^{(j)})$. For a new point, $\x^{(k+1)}$, we then have 
\[ \begin{bmatrix}\mathbf{f}_k\\f(\x^{(k+1)}) \end{bmatrix} \sim \normal(0, \begin{bmatrix} K & \mathbf{k}\\ \mathbf{k}^T & k(\x^{(k+1)}, \x^{(k+1)})
\end{bmatrix}  )  \]
with $\mathbf{k}_i = k(\x^{(k+1)}, \x^{(i)})$ for $i=1,...,k$.  From this, it can be shown that 
\[    
f(\x^{(k+1)})\mid \z_k, \x^{(k+1)} \sim \normal( \mathbf{k}^T K^{-1}\mathbf{f}_k,  k(\x^{(k+1)}, \x^{(k+1)})- 
 \mathbf{k}^T K^{-1}\mathbf{k})
   \]
So, once the kernel function is chosen, we have an explicit formula for the posterior distribution of $f$ with explicit predictive mean and covariance functions dependent upon prior observations and the kernel function. 

In addition to modelling $f$ as a random process, the Bayesian optimization paradigm uses an {\it acquisition function} (decision function) $a: \Omega \to \R$ to propose new inputs to sample.  The job of the acquisition function is to balance exploitation (pursuit of regions where the function is likely to be optimal) and exploration (generating draws in regions where $f$ may have a high degree of uncertainty) as well as to keep down the total number of evaluations of $f$. 

Once an acquisition function is selected, subsequent points are proposed via 
\[ \x_{k+1} = \argmax_{\x} a(\x \mid \z_k, \pmb{\theta})  \]
where, above, we use the notation $ a(\x \mid \z_k, \pmb{\theta})$ to denote the dependence of the acquisition function on the already sampled points and values together with any hyperparameters in the Gaussian Process representing $f$ (see \cite{practical_bayesian}).  When the Gaussian Process has a normally distributed posterior, popular acquisition functions have explicit dependence on the mean and variance of the posterior. 

Two acquisition functions worth mentioning are the {\it expected improvement} and {\it confidence bound}.  If $\x_+ = \argmax_{\x \in \x_k}f(\x)$ is the input for the best observed $f$ value thus far, then the improvement is\footnote{In the context of the CASH problem \eqref{CASH} the improvement is $\max(0, c(\blambda_-) -c(\blambda)))$ where $c$ is a given error rate dependent on conditional hierarchical hyperparameters $\blambda$ and $\blambda_-$ is the best performing currently observed hyperparameter selection.} 
\[ I(\x) = \max(0, f^{(k+1)}(\x) - f(\x_+))  \]
where $f^{(k+1)}(\x)$ is shorthand for $f(\x^{(k+1)})\mid \z_k, \x^{(k+1)}$, i.e. the current posterior for $f$.  While this has an explicit dependence on $f(\x)$, which we wish to avoid having to evaluate, the expected improvement acquisition function given by 
\[ a_{EI}(\x) = \expectation_{f\mid \x} I(\x)\]
where the expectation is over the distribution of $f\mid \z_k$, namely the current Gaussian process posterior.  This is expressible purely in terms of the predictive mean and variance of $f^{(k+1)}$ and $f(\x_+)$. 
The upper confidence bound acquisition function is given by 
\[ a_{UCB}(\x\mid \z_k, \pmb{\theta}) = \mu(\x\mid \z_k,\pmb{\theta}) + \kappa \sigma(\x\mid \z_k,\pmb{\theta}) \]
where $\mu, \sigma$ are the predictive distribution's mean and standard deviation, respectively.  In each of these acquisition functions, optimization to select the next point at which to sample $f$ is fairly straightforward. 

Putting the above considerations together, to perform a Gaussian process-based Bayesian optimization one repeatedly optimizes the acquisition function to generate new points to sample, then evaluates $f$ on the new point\footnote{Of course, in the machine learning hyperparameter context, the evaluation of $f$ may require training a model from scratch. 
}, then updates the posterior of the associated Gaussian Process model for $f$ to obtain newer predictive mean and variance.  This process repeats until a suitable optima is located within acceptable tolerances.

\subsection{Sequential Model-Based Optimizations}
\label{smac}
Given a dataset, let $c=c(\blambda)$ denote the loss function for a given machine learning model in terms of hyperparameters $\blambda$. We note that $\blambda$, as described in Section \ref{model_engineering} is hierarchical and conditional, and for this reason is tree-like.  
{\bf Sequential Model-Based Optimization} (SMBO) is a general optimization paradigm, similar to the Bayesian optimization approach described above, wherein the a surrogate to the objective function $f$ is optimized to find a proposed new point $\x^{(new)}$ to evaluate, upon which a new model for $f$ is constructed (see \cite{bergstra_algos}).  For hyperparameter optimization $c$ plays the role of $f$ and $\blambda$ plays the role of $\x$. We briefly discuss two approaches to the SMBO problem. 

{\bf Sequential Model-Based Algorithm Configuration} (SMAC, see \cite{smbo}) is an approach to SMBO using a variety of models for $c\mid \blambda$, including Gaussian processes and random forests.  In \cite{autoweka} the authors explore a SMAC paradigm using random forests.  Essentially, they model
\[c\mid \blambda \sim \normal(\mu_\blambda, \sigma^2_\blambda) \]
with $\mu_\blambda$ and $\sigma^2_\blambda$ determined by predictive average and variance across individual trees in a random forest. 
Under the assumption of normality for $c\mid \blambda$, SMAC offers a closed form expression for the expected improvement (see previous section) acquisition function. Additionally, SMAC enforces robustness by keeping alternating between hyperparameter configurations $\blambda^{(new)}$ suggested from the acquisition function, and those selected at random, in keeping with an exploitation/exploration paradigm.

A separate approach, discussed In \cite{bergstra_algos, empirical} and \cite{autoweka}, is using {\bf Tree-structured Parzen Estimators} (TPEs) to model the variables $\blambda \mid c$ and $\blambda$ to obtain $c\mid \blambda$.  In this approach
\[\prob( \blambda \mid c) = \ell(\blambda)\mathbbm{1}_{c<c^*}(c) + g(\blambda)\mathbbm{1}_{c\geq c^*}(c) \]
where $c^*$ is a quantile-based loss threshold and $\mathbbm{1}$ is an indicator variable and $\ell, g$ are empirical densities constructed on the associated observations. The density $\ell$, intuitively, should capture a distribution of well-performing hyperparameter values.  The surrogate expected improvement function then is proportional to $\frac{1}{\gamma + \frac{g(\blambda)}{\ell(\blambda)}(1-\gamma)}$, where $\gamma$ is the quantile defining $c^*$, thus leading to a simple optimization for the surrogate.

One-dimensional Parzen estimators\footnote{Parzen estimators are a standard technique for kernel density estimation.  See \cite{parzen}.} are then used on to estimate the density of each component of $\blambda$ during a tree traversal along paths whose nodes consist of active hyperparameters (i.e. components of $\blambda$).  The resultant one-dimensional estimates can then be combined to yield estimates of the multivariate densities $\ell(\blambda)$ and $g(\blambda)$, from which $c\mid \blambda$ can be estimated.

\end{document}